%% file: paper.tex
\documentclass{article}

\usepackage{microtype}
\usepackage{graphicx}
\usepackage{subcaption}
\usepackage{booktabs} %

\usepackage{hyperref}

\usepackage[preprint]{./configuration/icml2026}

\usepackage{amsmath}
\usepackage{amssymb}
\usepackage{mathtools}
\usepackage{amsthm}

\usepackage[capitalize, noabbrev]{cleveref}

\input{resources/preamble.tex}

\icmltitlerunning{Duality Models}

\begin{document}
\twocolumn[ \icmltitle{Duality Models: An Embarrassingly Simple One-step Generation Paradigm}

  \icmlsetsymbol{equal}{*}

  \begin{icmlauthorlist}
    \icmlauthor{Peng Sun}{wl,zju} \icmlauthor{Xinyi Shang}{ucl,mbzu}
    \icmlauthor{Tao Lin}{wl} \icmlauthor{Zhiqiang Shen}{mbzu}
  \end{icmlauthorlist}

  \icmlaffiliation{wl}{Westlake University, Hangzhou, China} \icmlaffiliation{zju}{Zhejiang University, Hangzhou, China} \icmlaffiliation{ucl}{University College London, London, UK} \icmlaffiliation{mbzu}{Mohamed bin Zayed University of Artificial Intelligence, Abu Dhabi, UAE}

  \icmlcorrespondingauthor{Tao Lin}{lintao@westlake.edu.cn}

  \icmlkeywords{Machine Learning, ICML}

  \vskip 0.3in ]

\printAffiliationsAndNotice{} %

\input{resources/main.tex}

\section*{Impact Statement}

This paper introduces Duality Models, a framework designed to significantly accelerate the inference of diffusion probabilistic models while maintaining high generation quality.
Our work has two primary societal implications:

\textbf{Environmental and Accessibility Impact:} By reducing the required sampling steps from hundreds to as few as one or two, our method substantially lowers the computational cost and energy consumption associated with generative model inference. This efficiency contributes to ``Green AI'' initiatives and helps democratize access to high-fidelity generation tools on consumer-grade hardware.

\textbf{Potential for Misuse:} As with any advancement in high-fidelity image synthesis, there is a risk that these techniques could be misused for creating misleading content or deepfakes. While our focus is on the fundamental optimization of the generative process, we advocate for the responsible deployment of such models, including the integration of watermarking and provenance tracking technologies to mitigate malicious use.

\bibliography{./resources/reference}
\bibliographystyle{./configuration/icml2026}

\input{resources/appendix.tex}
\end{document}

%% file: resources/preamble.tex
\usepackage{amsmath,amssymb,amsthm}

\providecommand{\norm}[1]{\left\lVert#1\right\rVert}

\providecommand{\E}{{\mathbb E}}
\providecommand{\E}[1]{{\mathbb E}\left.#1\right. }        %
\providecommand{\EEb}[2]{{\mathbb E}_{#1}\left[#2\right] } %

\providecommand{\dm}{\mathrm{d}}

\providecommand{\0}{\mathbf{0}}

\providecommand{\cc}{\mathbf{c}}

\providecommand{\uu}{\mathbf{u}}
\providecommand{\vv}{\mathbf{v}}

\providecommand{\xx}{\mathbf{x}}

\providecommand{\zz}{\mathbf{z}}

\providecommand{\mI}{\mathbf{I}}

\providecommand{\mtheta}{\boldsymbol{\theta}}

\providecommand{\mmF}{\boldsymbol{F}}

\providecommand{\cL}{\mathcal{L}}

\providecommand{\cN}{\mathcal{N}}

\newenvironment{talign*}
{\csname align*\endcsname}
{\endalign}

\usepackage[utf8]{inputenc}         %
\usepackage[T1]{fontenc}            %
\usepackage{url}                    %
\usepackage{booktabs}               %
\usepackage{amsfonts}               %
\usepackage{nicefrac}               %
\usepackage{microtype}              %
\usepackage{xcolor}                 %
\usepackage{algorithm}
\usepackage{algorithmic}
\usepackage{graphicx}
\usepackage{subcaption}
\usepackage[flushleft]{threeparttable}
\usepackage{float}
\usepackage{multirow}
\usepackage{xspace}
\usepackage{enumitem}
\usepackage[font=small]{caption}
\usepackage{autobreak}
\usepackage{sidecap}
\usepackage{wrapfig}
\usepackage{bbding}
\usepackage[toc, page, header]{appendix}
\usepackage{tikz}
\usepackage{xcolor}
\usepackage{pifont}
\usepackage{mdframed}
\usepackage{colortbl}

\hypersetup{
  colorlinks=true,
  linkcolor=blue,
  citecolor=teal,
  urlcolor=blue
}

\definecolor{coral}{RGB}{255,127,80}
\definecolor{darkgreen}{RGB}{0,100,0}
\definecolor{darkyellow}{RGB}{204,153,0}
\definecolor{salmon}{RGB}{250,128,114}
\definecolor{darkred}{RGB}{150,0,0}
\newcommand{\darkredtext}[1]{{\color{darkred}#1}}

\newcommand{\secref}[1]{\hyperref[#1]{\darkredtext{Sec.~\ref*{#1}}}}
\newcommand{\thmref}[1]{\hyperref[#1]{\darkredtext{Thm.~\ref*{#1}}}}
\newcommand{\defref}[1]{\hyperref[#1]{\darkredtext{Def.~\ref*{#1}}}}
\newcommand{\propref}[1]{\hyperref[#1]{\darkredtext{Prop.~\ref*{#1}}}}
\newcommand{\assumpref}[1]{\hyperref[#1]{\darkredtext{Assump.~\ref*{#1}}}}
\newcommand{\remarkref}[1]{\hyperref[#1]{\darkredtext{Rem.~\ref*{#1}}}}
\newcommand{\hypref}[1]{\hyperref[#1]{\darkredtext{Hyp.~\ref*{#1}}}}
\newcommand{\conjref}[1]{\hyperref[#1]{\darkredtext{Conj.~\ref*{#1}}}}
\newcommand{\lemref}[1]{\hyperref[#1]{\darkredtext{Lem.~\ref*{#1}}}}
\newcommand{\corref}[1]{\hyperref[#1]{\darkredtext{Cor.~\ref*{#1}}}}
\newcommand{\noteref}[1]{\hyperref[#1]{\darkredtext{Nota.~\ref*{#1}}}}
\newcommand{\claimref}[1]{\hyperref[#1]{\darkredtext{Clm.~\ref*{#1}}}}
\newcommand{\obsref}[1]{\hyperref[#1]{\darkredtext{Obs.~\ref*{#1}}}}
\newcommand{\algref}[1]{\hyperref[#1]{\darkredtext{Alg.~\ref*{#1}}}}
\newcommand{\figref}[1]{\hyperref[#1]{\darkredtext{Fig.~\ref*{#1}}}}
\newcommand{\tabref}[1]{\hyperref[#1]{\darkredtext{Tab.~\ref*{#1}}}}
\newcommand{\appref}[1]{\hyperref[#1]{\darkredtext{App.~\ref*{#1}}}}

\newtheoremstyle{custom}
{1pt} %
{1pt} %
{\itshape} %
{} %
{\bfseries} %
{} %
{ } %
{\thmname{#1} \thmnumber{#2} \thmnote{(#3)} . } %

\theoremstyle{custom}

\newtheorem{innerdefinition}{Definition}
\newtheorem{innerproposition}{Proposition}
\newtheorem{innerassumption}{Assumption}
\newtheorem{innerremark}{Remark}
\newtheorem{innertheorem}{Theorem}
\newtheorem{innerhypothesis}{Hypothesis}
\newtheorem{innerconjecture}{Conjecture}
\newtheorem{innerlemma}{Lemma}
\newtheorem{innercorollary}{Corollary}
\newtheorem{innernotation}{Notation}
\newtheorem{innerclaim}{Claim}
\newtheorem{innerproblem}{Problem}

\newtheorem{innerobservation}{Observation}

\newmdenv[
  backgroundcolor=gray!10,
  linecolor=gray!100,
  linewidth=0.8pt,
  skipabove=2pt,
  skipbelow=2pt,
  innertopmargin=5pt,
  innerbottommargin=5pt,
  innerleftmargin=5pt,
  innerrightmargin=5pt,
]{definitionframe}

\newmdenv[
  backgroundcolor=blue!10,
  linecolor=blue!100,
  linewidth=0.8pt,
  skipabove=2pt,
  skipbelow=2pt,
  innertopmargin=5pt,
  innerbottommargin=5pt,
  innerleftmargin=5pt,
  innerrightmargin=5pt,
]{propositionframe}

\newmdenv[
  backgroundcolor=green!10,
  linecolor=green!100,
  linewidth=0.8pt,
  skipabove=2pt,
  skipbelow=2pt,
  innertopmargin=5pt,
  innerbottommargin=5pt,
  innerleftmargin=5pt,
  innerrightmargin=5pt,
]{assumptionframe}

\newmdenv[
  backgroundcolor=yellow!10,
  linecolor=yellow!100,
  linewidth=0.8pt,
  skipabove=2pt,
  skipbelow=2pt,
  innertopmargin=5pt,
  innerbottommargin=5pt,
  innerleftmargin=5pt,
  innerrightmargin=5pt,
]{remarkframe}

\newmdenv[
  backgroundcolor=red!10,
  linecolor=red!100,
  linewidth=0.8pt,
  skipabove=7pt,
  skipbelow=1pt,
  innertopmargin=9pt,
  innerbottommargin=5pt,
  innerleftmargin=5pt,
  innerrightmargin=5pt,
]{theoremframe}

\newmdenv[
  backgroundcolor=purple!10,
  linecolor=purple!100,
  linewidth=0.8pt,
  skipabove=2pt,
  skipbelow=2pt,
  innertopmargin=5pt,
  innerbottommargin=5pt,
  innerleftmargin=5pt,
  innerrightmargin=5pt,
]{hypothesisframe}

\newmdenv[
  backgroundcolor=orange!10,
  linecolor=orange!100,
  linewidth=0.8pt,
  skipabove=2pt,
  skipbelow=2pt,
  innertopmargin=5pt,
  innerbottommargin=5pt,
  innerleftmargin=5pt,
  innerrightmargin=5pt,
]{conjectureframe}

\newmdenv[
  backgroundcolor=cyan!10,
  linecolor=cyan!100,
  linewidth=0.8pt,
  skipabove=2pt,
  skipbelow=2pt,
  innertopmargin=5pt,
  innerbottommargin=5pt,
  innerleftmargin=5pt,
  innerrightmargin=5pt,
]{lemmaframe}

\newmdenv[
  backgroundcolor=magenta!10,
  linecolor=magenta!100,
  linewidth=0.8pt,
  skipabove=2pt,
  skipbelow=2pt,
  innertopmargin=5pt,
  innerbottommargin=5pt,
  innerleftmargin=5pt,
  innerrightmargin=5pt,
]{corollaryframe}

\newmdenv[
  backgroundcolor=pink!10,
  linecolor=pink!100,
  linewidth=0.8pt,
  skipabove=2pt,
  skipbelow=2pt,
  innertopmargin=5pt,
  innerbottommargin=5pt,
  innerleftmargin=5pt,
  innerrightmargin=5pt,
]{notationframe}

\newmdenv[
  backgroundcolor=violet!10,
  linecolor=violet!100,
  linewidth=0.8pt,
  skipabove=2pt,
  skipbelow=2pt,
  innertopmargin=5pt,
  innerbottommargin=5pt,
  innerleftmargin=5pt,
  innerrightmargin=5pt,
]{claimframe}

\newmdenv[
  backgroundcolor=salmon!10,
  linecolor=salmon!100,
  linewidth=0.8pt,
  skipabove=2pt,
  skipbelow=2pt,
  innertopmargin=5pt,
  innerbottommargin=5pt,
  innerleftmargin=5pt,
  innerrightmargin=5pt,
]{problemframe}

\newmdenv[
  backgroundcolor=lavender!10,
  linecolor=lavender!100,
  linewidth=0.8pt,
  skipabove=2pt,
  skipbelow=2pt,
  innertopmargin=5pt,
  innerbottommargin=5pt,
  innerleftmargin=5pt,
  innerrightmargin=5pt,
]{observationframe}

\newenvironment{theorem}
{\begin{theoremframe}\begin{innertheorem}}
      {\end{innertheorem}\end{theoremframe}}

\usepackage[textwidth=2.6cm,textsize=tiny]{todonotes}
\definecolor{phasecolor}{RGB}{30, 100, 180} %
\newcommand{\algphase}[1]{\vspace{0.2em}\STATE \textbf{\textcolor{phasecolor}{#1}}\vspace{0.1em}}
\newcommand{\method}{\textsc{DuMo}\xspace} %

\newcommand{\sg}{\mathop{{\color{red}\operatorname{sg}}}\nolimits}

\newcommand{\rp}{$\boldsymbol{\color{red!70!black}\oplus}$}

\usepackage{algorithm}
\usepackage{listings}
\usepackage{xcolor}
\usepackage{caption}

\definecolor{codegreen}{rgb}{0.3, 0.5, 0.5}
\definecolor{codepink}{rgb}{0.85, 0.2, 0.5}
\definecolor{codeblue}{rgb}{0.2, 0.2, 1.0}

%% file: resources/main.tex
\begin{abstract}
    Consistency-based generative models like Shortcut and MeanFlow achieve impressive results via a target-aware design for solving the Probability Flow ODE (PF-ODE).
    Typically, such methods introduce a target time $r$ alongside the current time $t$ to modulate outputs between a local multi-step derivative ($r=t$) and a global few-step integral ($r=0$).
    However, the conventional ``one input, one output'' paradigm enforces a partition of the training budget, often allocating a significant portion (e.g., $75\%$ in MeanFlow) solely to the multi-step objective for stability.
    This separation forces a trade-off: allocating sufficient samples to the multi-step objective leaves the few-step generation undertrained, which harms convergence and limits scalability. \\
    To this end, we propose \textbf{Duality Models} (\method) via a ``one input, dual output'' paradigm.
    Using a shared backbone with dual heads, \method simultaneously predicts velocity $\vv_{t}$ and flow-map $\uu_{t}$ from a single input $\xx_{t}$.
    This applies geometric constraints from the multi-step objective to every sample, bounding the few-step estimation without separating training objectives, thereby significantly improving stability and efficiency.
    On ImageNet $256\times256$, a $679$M Diffusion Transformer with SD-VAE achieves a state-of-the-art (SOTA) \textbf{FID of $\mathbf{1.79}$ in just $\mathbf{2}$ steps}.
    Code is available at: \url{https://github.com/LINs-lab/DuMo}.
\end{abstract}

\section{Introduction}

\begin{figure}[t]
    \centering
    \includegraphics[width=\linewidth]{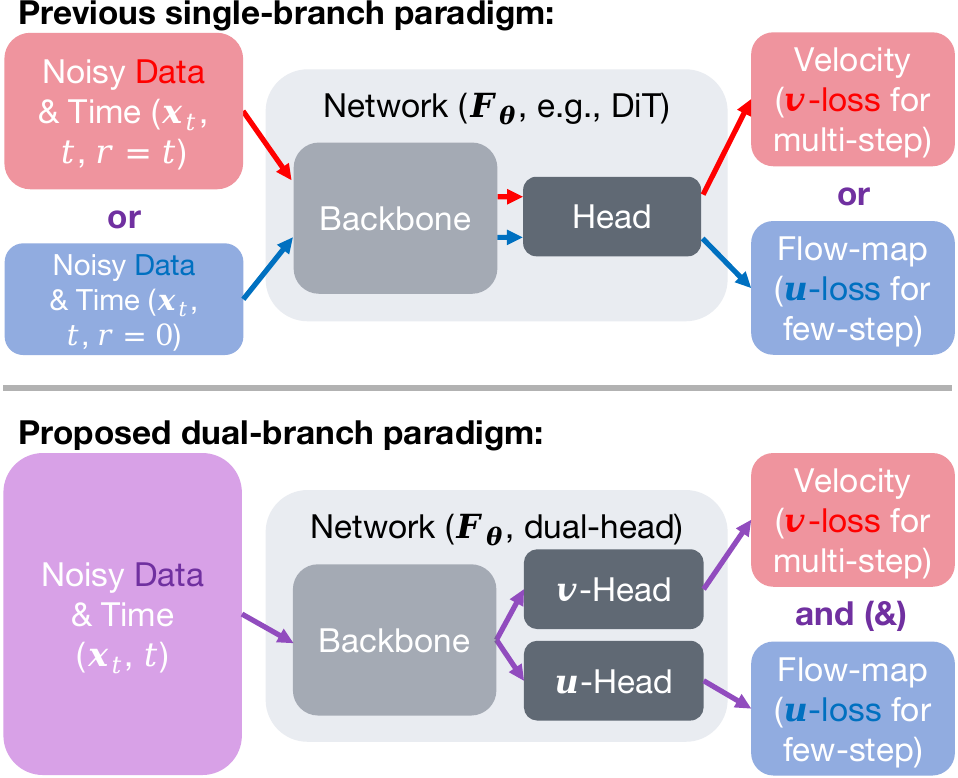}
    \vspace{-1.0em}
    \caption{\small
        \textbf{Schematic comparison.}
        (Top) Existing one-step models~\cite{geng2025mean,cheng2025twinflow} adopt a ``one input, one output'' scheme, predicting either velocity $\vv_{t}$ or flow-map $\uu_{t}$ conditioned on $r$.
        (Bottom) \method employs a ``one input, dual output'' design to simultaneously predict $\vv_{t}$ and $\uu_{t}$ from $\xx_{t}$.
        This incurs negligible overhead ($<\mathbf{0.5\%}$, e.g., +3M on a 675M DiT) by adding only an output head while preserving the backbone.
    }
    \label{fig:visual_demo}
    \vspace{-1em}
\end{figure}

Diffusion probabilistic models have fundamentally transformed the landscape of generative Artificial Intelligence, demonstrating unprecedented capability in high-fidelity image synthesis and distribution modeling~\cite{ho2020denoising, song2020score}.
Despite their success, the standard formulation requires solving the PF-ODE through dozens or hundreds of iterative steps, creating a significant computational bottleneck during inference.
To mitigate this latency, the community has actively explored acceleration techniques, with \textit{consistency-based methods}~\cite{song2023consistency} emerging as a premier paradigm for mapping noise to data in as few as one or two steps.

However, scaling consistency-based methods to large-scale, high-resolution models remains a formidable challenge due to their notorious training instability~\cite{sun2026anystep,cheng2025twinflow}.
Without sufficient local constraints, the recursive enforcement of self-consistency often leads to optimization divergence or mode collapse.
To counteract this volatility, recent SOTA approaches, such as Shortcut~\cite{frans2024one}, MeanFlow~\cite{geng2025mean} and RCGM~\cite{sun2026anystep}, have adopted a \textit{target-aware} design.
These methods introduce a target time $r$ to modulate the learning objective: the model minimizes a stable multi-step regression loss (velocity field) when $r \approx t$, and switches to the volatile few-step consistency loss (flow-map) only when $r=0$.
By anchoring the training process in the stable velocity field, these methods prevent divergence, enabling consistency training to scale to complex datasets.

While effective for stability, we contend that this ``one input, one output'' paradigm imposes a \textbf{structural bottleneck on the generation quality}.
Since the network is constrained to predict \textit{either} the velocity \textit{or} the flow-map for any given input, the training process necessitates a trade-off in data partitioning.
To prevent divergence, existing methods are compelled to route the vast majority of training samples (e.g., up to $75\%$ in MeanFlow~\cite{geng2025mean}) exclusively to the auxiliary velocity task.
Crucially, this leaves the flow-map head—which is solely responsible for inference—effectively \textbf{``starved'' of supervision}.
This severe data scarcity does not merely slow down convergence: it \textbf{fundamentally caps the performance ceiling}, trapping the one-step generator in a suboptimal regime despite the large-scale pre-training~\cite{geng2025mean,frans2024one,lu2024simplifying}.
Moreover, this issue becomes particularly acute when training large-scale models on massive datasets~\cite{sun2026anystep,cheng2025twinflow}.

In this paper, we propose to resolve this trade-off via a novel \textbf{Duality Models} (\method) framework.
Breaking away from the conditional switching paradigm, \method employs a simple yet effective ``one input, dual output'' architecture that simultaneously predicts both the local velocity $\vv_{t}$ and the global flow-map $\uu_{t}$ from a single noisy input $\xx_{t}$.
This design is mathematically elegant: it allows the geometric constraints of the PF-ODE (provided by the velocity head) to act as a regularizer for the flow-map prediction on \textit{every single training sample}, without the need to partition the training data budget.
By jointly optimizing these objectives in a single model with dual heads, \method inherently bridges the gap between stability and data efficiency.

\textbf{Our contributions are summarized as follows}:
\begin{enumerate}[label=(\alph{*}), nosep, leftmargin=16pt]
    \item We identify a critical bottleneck in existing target-aware models: the requisite partition of the training budget creates a data scarcity for the generative objective, limiting the performance ceiling of one-step sampling.

    \item We propose \method, a ``one input, dual output'' framework that unifies the learning of velocity and flow-maps.
          This design leverages multi-step physics to bound and regularize few-step estimation on every sample, ensuring both stability and efficiency.

    \item We achieve SOTA performance on ImageNet $256\times256$. \method enables a 679M-parameter Diffusion Transformer with SD-VAE to reach an FID of $1.79$ in just $2$ sampling steps, surpassing previous teacher-free methods in both convergence speed and generation quality.
\end{enumerate}

We detail our \textit{remarkably simple and effective} framework \method in \algref{alg:dumo}.
Notably, \method does \textit{not} require complex parameter design, specialized loss functions, or learning curricula to stabilize training and scaling, as justified by our comprehensive empirical validation in \secref{sec:exp}.

\begin{figure*}[t!]
    \centering
    \includegraphics[width=\linewidth]{
        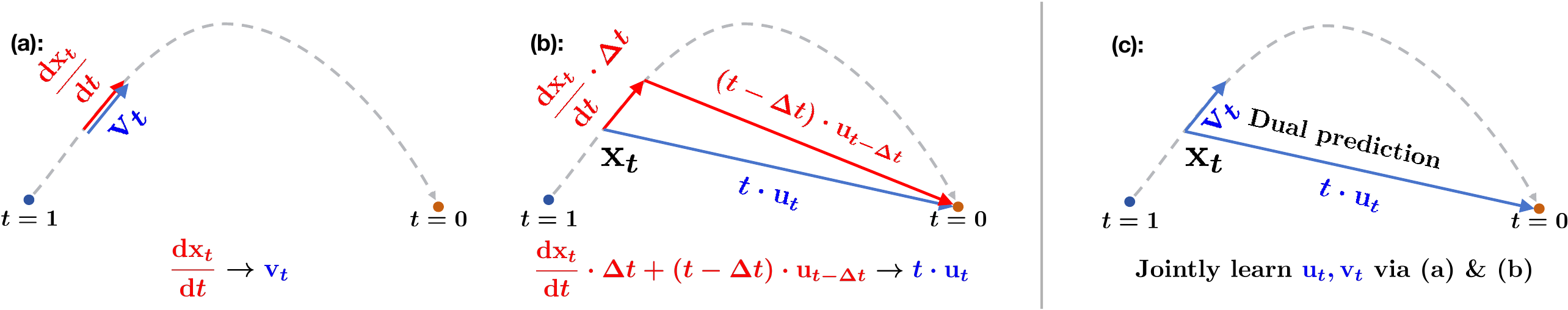
    }
    \caption{\small
        \textbf{Conceptual illustration of learning paradigms.}
        Trajectories illustrate the mapping from a \textcolor{blue}{current prediction state} to a \textcolor{red}{target learning state}.
        \textbf{(a)} Multi-step methods, such as standard diffusion~\citep{ho2020denoising} and flow-matching~\citep{lipman2022flow}, learn the local derivative (velocity).
        \textbf{(b)} Prominent one-step approaches, including consistency models~\citep{song2023consistency}, MeanFlow~\citep{geng2025mean}, and Shortcut models~\citep{frans2024one}, learn the global integral (flow-map).
        \textbf{(c)} \method unifies these paradigms via a dual-output architecture, establishing a natural combination of the geometric constraints from (a) and the generative efficiency of (b).
    }
    \vspace{-1.0em}
    \label{fig:visualization}
\end{figure*}

\section{Background and Preliminaries} \label{sec:preliminaries}
Generative modeling via diffusion aims to learn a mapping from a simple prior distribution $p_{1}(\zz) = \cN(\0, \mI)$ to a complex data distribution $p_{0}(\xx)$.
Modern continuous-time generative models, including Diffusion Models~\citep{ho2020denoising,song2020score} and Flow Matching (FM)~\citep{lipman2022flow}, unify this process through the lens of the PF-ODE. \looseness=-1

Consider a probability path $p_{t}(\xx)$ interpolating between data at $t=0$ and noise at $t=1$.
A sample $\xx_{t}$ along this path is typically constructed via an affine transformation of data $\xx \sim p_{0}$ and noise $\zz \sim p_{1}$:
\begin{equation}
    \xx_{t} = \alpha(t)\zz + \gamma(t)\xx \,, \label{eq:forward_process}
\end{equation}
where $\alpha(t), \gamma(t) \in C^{1}[0,1]$ are schedule functions satisfying boundary conditions $\alpha(0)=0, \gamma(0)=1$ (pure data) and $\alpha(1)=1, \gamma(1)=0$ (pure noise).
The trajectory of samples over time defines a vector field $\vv_{t}(\xx_{t})$, which characterizes the instantaneous velocity of the probability mass transport.
The corresponding PF-ODE is given by:
\begin{equation} \label{eq:pf_ode}
    \frac{\mathrm{d}\xx_{t}}{\mathrm{d}t}= \vv_{t}(\xx_{t}) \,, \quad
    \text{where }\vv_{t}(\xx_{t}) \text{ is the velocity field}.
\end{equation}
Generative sampling is performed by numerically integrating~\eqref{eq:pf_ode} from $t=1$ to $t=0$.
Depending on the learning target relative to this ODE, existing methods can be broadly categorized into two paradigms: learning the local derivative (Multi-step) or learning the global integral (Few-step).

\subsection{Multi-step Models: Learning the Local Derivative} \label{sec:prelim_multistep}
Diffusion and Flow-matching models focus on learning the local geometry of the PF-ODE.
Specifically, they aim to approximate the velocity field $\vv_{t}$ defined in~\eqref{eq:pf_ode}.
As unified in previous studies~\citep{sun2025unified,lipman2022flow}, despite differences in transport paths, the core objective is to regress a neural network $\vv_{\mtheta}( \xx_{t}, t)$ to the ground-truth conditional velocity field.
The general loss function is:
\begin{equation} \label{eq:flow_matching_loss}
    \cL_{\vv}(\mtheta) = \EEb{t, \xx, \zz}{\norm{\vv_{\mtheta}(\xx_t, t) - \dot{\xx}_t}_2^2} \,,
\end{equation}
where $\dot{\xx}_{t}= \frac{\mathrm{d}}{\mathrm{d}t}(\alpha(t)\zz + \gamma(t)\xx)$ represents the target velocity derived from the forward process.
A conceptual illustration of this learning paradigm is provided in \figref{fig:visualization}(a).

\paragraph{Remarks.}
This paradigm relies on a simple regression objective ($\ell_{2}$ loss) to a stable target.
Consequently, training is notoriously stable and scalable.
However, sampling requires solving the ODE with small step sizes to faithfully follow the local curvature, necessitating dozens or hundreds of function evaluations (NFE), which hinders real-time applications. \looseness=-1

\subsection{Consistency-based Few-step Models}
To accelerate sampling, Consistency Models~\citep{song2023consistency} and Flow-map models~\citep{geng2025mean,frans2024one} learn the PF-ODE integral directly.
Unlike local derivative methods ($\vv_{t}$), these approaches parameterize a \textit{global flow-map} $\uu_{\mtheta}(\xx_{t}, t)$.
Here, $t \cdot \uu_{\mtheta}(\xx_{t}, t)$ represents a straight-line vector from $\xx_{t}$ to the data $\xx_{0}$.

The \textbf{self-consistency principle} behind consistency models states that points on the same ODE trajectory must map to the same origin.
Consistency models enforce this by ensuring predictions at $t$ and $t-\Delta t$ remain consistent:
\begin{equation} \label{eq:consistency_condition}
    t \cdot \uu_{\mtheta}(\xx_{t}, t) \equiv \frac{\mathrm{d}\mathbf{x}_{t}}{\mathrm{d}t} \cdot \Delta t + (t -\Delta t) \uu_{\mtheta}(\xx_{t-\Delta t}, t)  \,.
\end{equation}
Training uses a recursive objective where the model learns from its own predictions (bootstrapping) rather than fixed ground truth.
Recent methods~\citep{frans2024one,geng2025mean,sun2026anystep} introduce a target time $r$ to modulate between local and global predictions; see \figref{fig:visualization}(b).

\paragraph{Remarks.}
By approximating the integral mapping, these models facilitate 1-step or few-step generation.
However, training stability is often compromised by the recursive nature of the objective (i.e., the ``moving target'' problem) and the difficulty of learning the global flow without local constraints.
Consequently, these methods often require complex training schedules or large batch sizes, presenting a trade-off between training stability and sampling efficiency.

\section{Methodology}
\label{sec:methodology}

\begin{figure*}[t!]
    \centering
    \includegraphics[width=\linewidth]{
        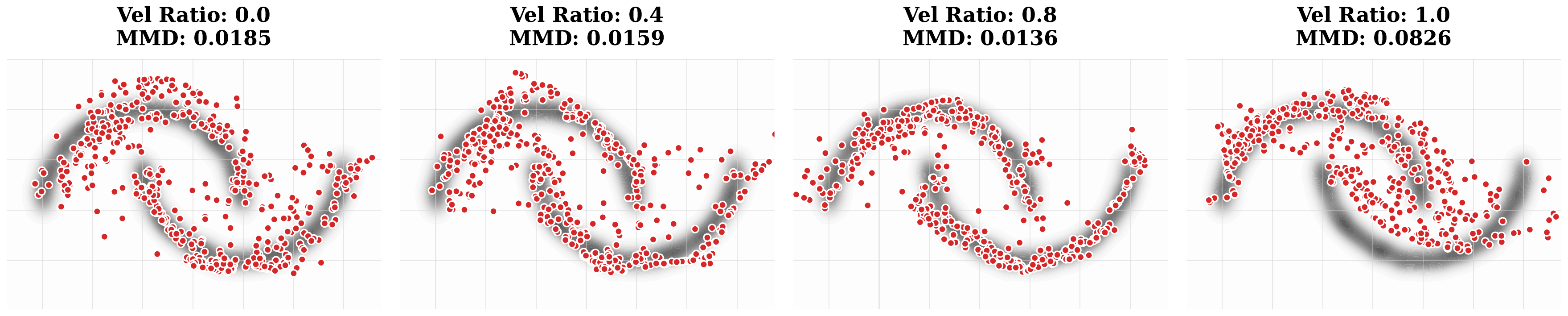
    }
    \caption{\small
        \textbf{Impact of Velocity Ratio ($\rho$) on one-step generation using MeanFlow~\citep{geng2025mean}.}
        We visualize samples generated on the Moons dataset to analyze the behavior of this representative single-branch model.
        \textcolor{gray}{Gray areas}: Ground truth distribution;
        \textcolor{red}{Red dots}: Samples generated via one-step inference.
        We report Maximum Mean Discrepancy (MMD) to quantify quality (lower is better).
        The results highlight a rigid trade-off: low velocity supervision ($\rho=0$) leads to instability/divergence, while excessive supervision ($\rho=1.0$) hampers the learning of the few-step mapping.
        Optimal performance is achieved at $\rho=0.8$, confirming that MeanFlow requires a specific partition of the training budget to balance stability and efficacy.
    }
    \vspace{-1.0em}
    \label{fig:moons_experiment}
\end{figure*}

In this section, we present \method, a framework designed to overcome the structural limitations of existing target-aware generative models.
We begin by analyzing the dilemma inherent in the single-branch paradigm, identifying a critical trade-off between training stability and sample efficiency (\secref{sec:revisit}).
Building on this analysis, we introduce our \textbf{Dual-Branch} design, which unifies the learning of local velocity and global flow-maps to resolve this conflict (\secref{sec:dual_design}).
Finally, we detail the practical implementation of our framework, including the architectural modifications and the training algorithm (\secref{sec:practical} and \algref{alg:dumo}).

\subsection{Revisiting the Dilemma in Single-Branch Paradigm} \label{sec:revisit}
Modern SOTA one-step generative models, such as MeanFlow~\citep{geng2025mean} and Shortcut models~\citep{frans2024one}, predominantly adopt a \textit{single-branch paradigm} coupled with a mixture optimization strategy.
Formally, these methods can be viewed as optimizing a composite objective governed by a hyperparameter $\rho$ (the Velocity Ratio):
\begin{equation}
    \cL(\mtheta) = \E_{\xx, \zz, t, p}[ \mathbb{I}_{p < \rho}\cdot \underbrace{\mathcal{L}_{\vv}(\mtheta)}_{\text{\textcolor{blue}{Stability}}}+ \mathbb{I}_{p \ge \rho}\cdot \underbrace{\mathcal{L}_{\uu}(\mtheta)}_{\text{\textcolor{red}{Efficacy}}}] \, ,
\end{equation}
where $p \sim \text{Uniform}(0, 1)$ is a random variable determining the task for the current step, and $\cL_{\vv}, \cL_{\uu}$ denote the multi-step (velocity) and few-step (flow-map) losses, respectively, as defined in \secref{sec:preliminaries}.
This formulation creates a \textbf{zero-sum trade-off} between two critical requirements:
\begin{enumerate}[label=(\alph{*}), nosep, leftmargin=16pt]
    \item \textbf{Training stability:}
          The velocity loss $\mathcal{L}_{\vv}$ is indispensable for grounding the generation process in the physical constraints of the PF-ODE.
          As evidenced in prior studies~\citep{sun2025unified,geng2025mean,sun2026anystep,cheng2025twinflow}, this term provides stable, low-variance supervision.
          We illustrate this empirically in \figref{fig:moons_experiment}: removing this supervision ($\rho=0.0$) leads to training divergence (high MMD).
          Consequently, existing methods are forced to allocate a dominant portion of the training budget (e.g., $\rho \in [0.75, 0.9]$) solely to stabilize the optimization~\cite{geng2025mean}.

    \item \textbf{Few-step generation efficacy:}
          Conversely, the capability for one-step generation stems directly from $\mathcal{L}_{\uu}$.
          However, the high $\rho$ required for stability inevitably limits the effective training data for this objective.
          As shown in \figref{fig:moons_experiment}, while a balanced ratio improves results, an excessive focus on stability ($\rho=1.0$) fails to optimize the flow-map effectively, thus significantly degrading generation quality.
\end{enumerate}

In practice, this dilemma forces a compromise: models must sacrifice data efficiency to prevent divergence.
The root cause lies in the \textbf{single-branch constraint}: the network is structurally incapable of leveraging the geometric stability of the ODE derivative to regularize the flow-map prediction within a single forward pass.
This observation motivates our proposed Dual-Branch design, which aims to dissolve this trade-off by simultaneously satisfying both objectives.

\subsection{Key Design of Dual-Branch Paradigm} \label{sec:dual_design}
In this section, we elaborate on the architectural design of \method and justify its efficacy.
Without loss of generality, we anchor our exposition in the ``Linear'' transport path (i.e., $\alpha(t) = t, \gamma (t) = 1-t$), a setting widely adopted in recent one-step generative models~\citep{frans2024one,geng2025mean,sun2026anystep,sun2025unified,cheng2025twinflow}.

\paragraph{Dual-branch diffusion and flow models.}
Conventional single-branch diffusion and flow-matching models typically parameterize a network $\mmF_{\mtheta}(\xx_{t}, t)$ to approximate the velocity field alone~\cite{sun2025unified,ho2020denoising}.
In contrast, we propose a \textit{dual-branch} formulation where the model simultaneously predicts two distinct geometric quantities:
\begin{equation}
    \mmF_{\mtheta}(\xx_{t}, t) = (\vv_{\mtheta,t}, \, \uu_{\mtheta,t}) \, ,
\end{equation}
which aim to learn the instantaneous velocity $\vv_{t}:= \frac{\mathrm{d}\xx_{t}}{\mathrm{d} t}$ and the flow-map $\uu_{t}:= \frac{\xx_{t}- \xx_{0}}{t}$ targeting the data endpoint $\xx_{0}$ along the PF-ODE trajectory passing through $\xx_{t}$.

\paragraph{Dual loss design in \method.}
Leveraging this simultaneous prediction capability, we construct a composite objective that unifies local geometric constraints with global consistency enforcement.
For the velocity head, we adopt the standard flow-matching regression loss:
\begin{equation}
    \label{eq:v_head_loss}\cL_{\vv}(\mtheta) = \EEb{\xx_0, \zz, t}{ \norm{ \vv_{\mtheta,t} - \vv_t}_2^2 } \, ,
\end{equation}
where the target velocity is given by $\vv_{t}= \zz - \xx_{0}$ under the linear path.
For the flow-map head, we employ the consistency training objective~\cite{song2023consistency,lu2024simplifying,geng2025mean,sun2026anystep}:
\begin{equation}
    \label{eq:u_head_loss}\cL_{\uu}(\mtheta) = \EEb{\xx_0, \zz, t}{ \norm{ \uu_{\mtheta,t} - \vv_t + t \cdot \frac{\mathrm{d}\uu_{\mtheta^-,t}}{\mathrm{d} t}}_2^2 } \, ,
\end{equation}
where $\mtheta^{-}$ denotes a copy of the parameters with stopped gradients (or a target network).
The total training objective is a weighted combination of these two terms:
\begin{equation}
    \label{eq:final_loss}\cL(\mtheta) = \beta \cdot \cL_{\vv}(\mtheta) + (1-\beta) \cdot \cL_{\uu}(\mtheta) \, .
\end{equation}
A conceptual illustration of this learning paradigm is provided in \figref{fig:visualization}(c).
Intuitively, the hyperparameter $\beta$ governs the trade-off between \textit{training stability}—inherent to the velocity regression $\cL_{\vv}$—and \textit{generation efficiency}, which is driven by the consistency objective $\cL_{\uu}$.
We provide a detailed analysis of $\beta$ and its empirical settings in \secref{sec:exp}.

\paragraph{Justification on compatibility.}
For \eqref{eq:final_loss}, one might question whether combining velocity and flow-map losses creates conflicting gradients, as these tasks are conventionally treated separately.
In \appref{app:theoretical_analysis}, \textit{we analyze and justify the compatibility of these objectives}, demonstrating that they are theoretically consistent within our framework.

\subsection{Practical Design Implementation}
\label{sec:practical}
This section details implementation strategies ensuring the stability and efficiency of \method. We cover time sampling, guidance integration, loss function selection, and architectural modifications for the dual-head paradigm.

\paragraph{Time distribution sampling.}
To align the training process with the signal-to-noise ratio of the data, we adopt a flexible time sampling strategy, following established practices~\citep{frans2024one,sun2025unified,sun2026anystep,song2023consistency}.
Specifically, time steps are sampled as $t \sim \mathrm{Beta}(\theta_{1}, \theta_{2})$.
The hyperparameters $\theta_{1}$ and $\theta_{2}$ allow us to bias the training focus towards specific noise levels (e.g., concentrating on mid-trajectory regions where consistency learning is most critical).
We provide detailed ablation studies on the choice of $\theta_{1}$ and $\theta_{2}$ in \secref{sec:ablation}.

\paragraph{Enhanced target velocity via guidance.}
It is well-established that incorporating guidance during training or sampling significantly improves perceptual quality~\citep{ho2022classifier,dhariwal2021diffusion,karras2022elucidating}.
Theoretically, this corresponds to modifying the conditional target score function from $\nabla_{\mathbf{x}_t}\log p_{t}(\mathbf{x}_{t})$ to an enhanced version $\nabla_{\xx_t}\log\left(p_{t}(\xx_{t}|\cc) \cdot \left(\frac{p_{t,\mtheta}(\xx_{t}|\cc)}{p_{t,\mtheta}(\xx_{t}|\varnothing)} \right)^{\zeta}\right)$, where $\zeta \in (0,1)$ denotes the enhancement ratio (guidance scale).

In practice, we implement this via \textit{Training-time Classifier-Free Guidance} (CFG).
We update the ground-truth velocity target $\vv_{t}$ used in \eqref{eq:v_head_loss} and \eqref{eq:u_head_loss} using the model's own predictions:
$\vv_{t}\leftarrow \vv_{t}+ \zeta \cdot (\vv_{\mtheta-,t}- \vv_{\mtheta^-,t, \varnothing})$, where $\vv_{\mtheta^-,t, \varnothing}$ represents the unconditional prediction from the EMA (or stop-gradient) model.
This aligns our implementation with recent SOTA frameworks~\citep{frans2024one,geng2025mean,sun2025unified,sun2026anystep}.

\paragraph{Simplicity in loss design.}
While the standard $\ell_{2}$-norm is ubiquitous in multi-step diffusion training~\citep{ho2020denoising}, few-step distillation methods often require complex, task-specific loss functions (e.g., LPIPS or heavily weighted $\ell_{2}$ variants) to prevent signal decay~\citep{song2023improved,geng2025mean}.
These specialized losses introduce additional hyperparameters (e.g., norm orders $p$) that necessitate extensive tuning for different resolutions or architectures.
\textit{
    In contrast, we demonstrate the \textbf{robustness and simplicity} of \method by utilizing the standard, unweighted $\ell_{2}$-norm for all experiments.
    Our method achieves SOTA results without relying on complex metric engineering.
}

\paragraph{Efficient $\uu$-loss calculation.}
The consistency loss term in \eqref{eq:u_head_loss} involves a time derivative $\frac{\mathrm{d}\uu_{\mtheta^-,t}}{\mathrm{d} t}$, which technically requires a Jacobian-vector product (JVP) computation~\citep{lu2024simplifying}.
To mitigate the computational overhead of JVP, we adopt the efficient \texttt{JVP}-free approximation proposed in~\citet{sun2025unified}.
This method utilizes a central finite-difference approximation to estimate the tangent vector, reducing the cost to simple forward passes without auto-differentiation through the graph.

\paragraph{Architectural adaptation for dual heads.}
We adapt standard single-head architectures (e.g., Diffusion Transformers~\cite{peebles2023scalable}) to the \method paradigm with minimal structural changes. We retain the entire backbone encoder and simply introduce a parallel output projection layer (head) for the flow-map $\uu_{t}$, while the original head continues to predict $\vv_{t}$.
The parameter overhead is negligible; for instance, adding a second head to a $675$M parameter DiT increases the model size by only $\sim 3$M parameters ($<0.5\%$). \looseness=-1

\paragraph{Computational efficiency.}
\algref{alg:dumo} summarizes the complete training procedure of \method.
A key advantage is that simultaneous prediction of $\vv_{\mtheta,t}$ and $\uu_{\mtheta,t}$ eliminates redundant forward passes during target enhancement.
Specifically, when computing the enhanced target velocity $\vv_{t}\leftarrow \vv_{t}+ \zeta \cdot (\vv_{\mtheta^-,t}- \vv_{\mtheta^-,t, \varnothing})$, standard methods require a separate forward pass to obtain $\vv_{\mtheta,t}$.
In our framework, this term is already available from the main forward pass (detached via \texttt{stopgrad}), eliminating redundant computation.

\textit{%
    \method achieves superior computational efficiency.
    Each training step requires only \textbf{1 gradient-tracking} and \textbf{3 gradient-free} forward passes.
    In comparison, SOTA single-branch methods like MeanFlow~\cite{geng2025mean} (using JVP-free implementations) require \textbf{1 gradient-tracking} and \textbf{4 gradient-free} passes, making \method approximately \textbf{12\% more efficient} per iteration (cf., \secref{sec:ablation} for more details).
}

\begin{algorithm}[t]
    \caption{\method Training Step}
    \label{alg:dumo}
    \begin{algorithmic}
        [1] \REQUIRE Data distribution $p(\mathbf{x})$, model parameters $\mtheta$, stop gradient operator $\sg$, enhancement ratio $\zeta$, Beta distribution parameters $(\theta_{1},\theta_{2})$, loss weighting factor $\beta$.

        \algphase{Preparation Phase:}
        \STATE Sample data $\mathbf{x}\sim p(\mathbf{x})$ and noise $\mathbf{z}\sim \mathcal{N}(\mathbf{0}, \mathbf{I})$.
        \STATE Sample time step $t \sim \mathrm{Beta}(\theta_{1}, \theta_{2})$.
        \STATE Construct perturbed state $\mathbf{x}_{t}= t \cdot \mathbf{z}+ (1 - t) \cdot \mathbf{x}$.
        \STATE Compute ground truth velocity $\mathbf{v}_{t}= \mathbf{z}- \mathbf{x}$.

        \algphase{Learning Target Estimation:}
        \STATE Compute model outputs $(\vv_{\mtheta,t}, \, \uu_{\mtheta,t}) = \mmF_{\mtheta}(\xx_{t}, t)$.
        \STATE Compute unconditional prediction $\vv_{\mtheta^-,t, \varnothing}$.
        \STATE Enhance velocity: $\vv_{t}\gets \vv_{t}+ \zeta \cdot (\sg(\vv_{\mtheta,t}) - \vv_{\mtheta^-,t, \varnothing})$.
        \STATE Compute flow-map target: $\uu_{t}= \vv_{t}- t \cdot \frac{\dm \uu_{\mtheta^-,t}}{\dm t}$.

        \algphase{Loss Calculation \& Model Update:}
        \STATE Update parameters $\mtheta$ by minimizing the objective:
        \[
            \mathcal{L}(\mtheta) = \beta \cdot \norm{ \vv_{\mtheta,t} - \vv_t}_{2}^{2}+ (1-\beta) \cdot \norm{ \uu_{\mtheta,t} - \uu_t}_{2}^{2}
        \]
    \end{algorithmic}
\end{algorithm}

\section{Experiments}
\label{sec:exp}

We empirically validate the \method framework.
We first detail the experimental setup, covering datasets, architectures, and training protocols, followed by an evaluation of generation quality and efficiency against SOTA baselines.

\subsection{Experimental Setup} \label{sec:expset}
\paragraph{Datasets and latent space operations.}
We conduct our primary evaluation on the ImageNet-1K dataset~\citep{deng2009imagenet} at resolutions of $256 \times 256$ and $512 \times 512$, adhering to established benchmarks in high-fidelity generative modeling~\citep{karras2024analyzing,song2023consistency}.
We follow the standard preprocessing pipeline from ADM~\citep{dhariwal2021diffusion}.
To ensure computational efficiency, all experiments are performed in the latent space:
\begin{enumerate}[label=(\alph{*}), nosep, leftmargin=16pt]
    \item \textbf{$256 \times 256$:} We utilize the widely adopted SD-VAE~\citep{rombach2022high} as the default autoencoder.

    \item \textbf{$512 \times 512$:} In addition to SD-VAE, we employ DC-AE~\citep{chen2024deep}, which offers a higher compression ratio (\textit{f32c32}), to mitigate the computational demands of high-resolution training.
\end{enumerate}

\paragraph{Network architectures.}
We adopt the Diffusion Transformer (DiT)~\citep{peebles2023scalable} as our backbone, consistent with recent SOTA few-step generation methods~\citep{frans2024one,geng2025mean,sun2025unified}.
Specifically, we use the DiT-XL/2 variant ($675$M parameters).
To adapt this single-branch architecture for our \method paradigm, we introduce a parallel output head for flow-map prediction ($\uu_{t}$) alongside the original velocity head ($\vv_{t}$).
Crucially, this modification incurs a negligible parameter overhead of approximately $3$M ($<0.5\%$ increase), preserving the model's computational efficiency.

\paragraph{Implementation details.}
Models are implemented in PyTorch~\citep{paszke2019pytorch} and trained using the AdamW optimizer~\citep{loshchilov2017decoupled} with $\beta_{1}=0.9, \beta_{2}=0.95$, and no weight decay.
We use a constant learning rate of $1 \times 10^{-4}$ and a global batch size of $512$.
Further hyperparameter configurations are analyzed and detailed in \secref{sec:ablation} and \appref{app:implementation_details}.
Evaluation is performed using the standard Fr\'echet Inception Distance (FID)~\citep{heusel2017gans}, computed on $50,000$ generated samples against the full training set, following the protocol of~\citet{ho2020denoising}.

\begin{table*}[!t]
    \centering
    \caption{\small{\textbf{System-level quality comparison for few-step generation task on class-conditional ImageNet-1K.}}
        The \textbf{best} results of each resolution are highlighted.
        Notation A\rp B denotes the result obtained by combining methods A and B.}
    \vspace{-0.5em}
    \label{tab:in512and256_fewsteps}
    \resizebox{\textwidth}{!}{%
        \begin{tabular}{lcccc|lcccc}
            \toprule \multicolumn{5}{c|}{$512\times512$}                   & \multicolumn{5}{c}{$256\times256$}                                                                                                                                                                                                                                                          \\
            \cmidrule(lr){1-5} \cmidrule(lr){6-10} \textbf{Method}         & \textbf{NFE~$\downarrow$}          & \textbf{FID~$\downarrow$} & \textbf{\#Params}      & \textbf{\#Epochs}      & \textbf{Method}                                           & \textbf{NFE~$\downarrow$}      & \textbf{FID~$\downarrow$} & \textbf{\#Params}      & \textbf{\#Epochs}      \\
            \midrule \multicolumn{10}{c}{\textcolor{gray}{\textbf{Diffusion \& flow-matching models}}}                                                                                                                                                                                                                                                                   \\
            \midrule \textcolor{gray}{ADM-G~\citep{dhariwal2021diffusion}} & \textcolor{gray}{250$\times$2}     & \textcolor{gray}{7.72}    & \textcolor{gray}{559M} & \textcolor{gray}{388}  & \textcolor{gray}{ADM-G~\citep{dhariwal2021diffusion}}     & \textcolor{gray}{250$\times$2} & \textcolor{gray}{4.59}    & \textcolor{gray}{559M} & \textcolor{gray}{396}  \\
            \textcolor{gray}{U-ViT-H/4~\citep{bao2023all}}                 & \textcolor{gray}{50$\times$2}      & \textcolor{gray}{4.05}    & \textcolor{gray}{501M} & \textcolor{gray}{400}  & \textcolor{gray}{U-ViT-H/2~\citep{bao2023all}}            & \textcolor{gray}{50$\times$2}  & \textcolor{gray}{2.29}    & \textcolor{gray}{501M} & \textcolor{gray}{400}  \\
            \textcolor{gray}{DiT-XL/2~\citep{peebles2023scalable}}         & \textcolor{gray}{250$\times$2}     & \textcolor{gray}{3.04}    & \textcolor{gray}{675M} & \textcolor{gray}{600}  & \textcolor{gray}{DiT-XL/2~\citep{peebles2023scalable}}    & \textcolor{gray}{250$\times$2} & \textcolor{gray}{2.27}    & \textcolor{gray}{675M} & \textcolor{gray}{1400} \\
            \textcolor{gray}{SiT-XL/2~\citep{ma2024sit}}                   & \textcolor{gray}{250$\times$2}     & \textcolor{gray}{2.62}    & \textcolor{gray}{675M} & \textcolor{gray}{600}  & \textcolor{gray}{SiT-XL/2~\citep{ma2024sit}}              & \textcolor{gray}{250$\times$2} & \textcolor{gray}{2.06}    & \textcolor{gray}{675M} & \textcolor{gray}{1400} \\
            \textcolor{gray}{MaskDiT~\citep{zheng2023fast}}                & \textcolor{gray}{79$\times$2}      & \textcolor{gray}{2.50}    & \textcolor{gray}{736M} & \textcolor{gray}{-}    & \textcolor{gray}{DDT-XL/2~\citep{wang2025ddt}}            & \textcolor{gray}{250$\times$2} & \textcolor{gray}{1.26}    & \textcolor{gray}{675M} & \textcolor{gray}{400}  \\
            \textcolor{gray}{EDM2-S~\citep{karras2024analyzing}}           & \textcolor{gray}{63}               & \textcolor{gray}{2.56}    & \textcolor{gray}{280M} & \textcolor{gray}{1678} & \textcolor{gray}{REPA-XL/2~\citep{yu2024representation}}  & \textcolor{gray}{250$\times$2} & \textcolor{gray}{1.96}    & \textcolor{gray}{675M} & \textcolor{gray}{200}  \\
            \textcolor{gray}{EDM2-L~\citep{karras2024analyzing}}           & \textcolor{gray}{63}               & \textcolor{gray}{2.06}    & \textcolor{gray}{778M} & \textcolor{gray}{1476} & \textcolor{gray}{REPA-XL/2~\citep{yu2024representation}}  & \textcolor{gray}{250$\times$2} & \textcolor{gray}{1.42}    & \textcolor{gray}{675M} & \textcolor{gray}{800}  \\
            \textcolor{gray}{EDM2-XXL~\citep{karras2024analyzing}}         & \textcolor{gray}{63}               & \textcolor{gray}{1.91}    & \textcolor{gray}{1.5B} & \textcolor{gray}{734}  & \textcolor{gray}{Light.DiT~\citep{yao2025reconstruction}} & \textcolor{gray}{250$\times$2} & \textcolor{gray}{2.11}    & \textcolor{gray}{675M} & \textcolor{gray}{64}   \\
            \textcolor{gray}{DiT-XL\rp DC-AE}                              & \textcolor{gray}{250$\times$2}     & \textcolor{gray}{2.41}    & \textcolor{gray}{675M} & \textcolor{gray}{400}  & \textcolor{gray}{Light.DiT~\citep{yao2025reconstruction}} & \textcolor{gray}{250$\times$2} & \textcolor{gray}{1.35}    & \textcolor{gray}{675M} & \textcolor{gray}{800}  \\
            \midrule \multicolumn{10}{c}{\textbf{GANs}}                                                                                                                                                                                                                                                                                                                  \\
            \midrule BigGAN~\citep{brock2018large}                         & 1                                  & 8.43                      & 160M                   & -                      & BigGAN~\citep{brock2018large}                             & 1                              & 6.95                      & 112M                   & -                      \\
            StyleGAN~\citep{sauer2022stylegan}                             & 1$\times$2                         & \textbf{2.41}             & 168M                   & -                      & GigaGAN~\citep{kang2023scaling}                           & 1                              & \textbf{3.45}             & 569M                   & -                      \\
            \midrule \multicolumn{10}{c}{\textbf{Masked \& autoregressive models}}                                                                                                                                                                                                                                                                                       \\
            \midrule                                                                                   %
            MaskGIT~\citep{chang2022maskgit}                               & 12                                 & 7.32                      & 227M                   & 300                    & MaskGIT~\citep{chang2022maskgit}                          & 8                              & 6.18                      & 227M                   & 300                    \\
            VAR-$d$36-s~\citep{tian2024visual}                             & 10$\times$2                        & \textbf{2.63}             & 2.3B                   & 350                    & VAR-$d$30-re~\citep{tian2024visual}                       & 10$\times$2                    & \textbf{1.73}             & 2.0B                   & 350                    \\
            \midrule \multicolumn{10}{c}{\textbf{Single-branch one-step models (distillation-based)}}                                                                                                                                                                                                                                                                    \\
            \midrule sCD-M~\citep{lu2024simplifying}                       & 1                                  & 2.75                      & 498M                   & 1997                   & UCGM-XL/2~\citep{sun2025unified}                          & 1                              & 2.10                      & 675M                   & 424                    \\
                                                                           & 2                                  & 2.26                      & 498M                   & 1997                   &                                                           & 2                              & 1.94                      & 675M                   & 424                    \\
            sCD-L~\citep{lu2024simplifying}                                & 1                                  & 2.55                      & 778M                   & 1434                   & UCGM\rp REPA-XL/2                                         & 2                              & 1.95                      & 675M                   & 801                    \\
                                                                           & 2                                  & 2.04                      & 778M                   & 1434                   & UCGM\rp Light.DiT                                         & 2                              & 2.06                      & 675M                   & 801                    \\
            sCD-XXL~\citep{lu2024simplifying}                              & 1                                  & 2.28                      & 1.5B                   & 921                    & UCGM\rp DDT-XL/2                                          & 2                              & \textbf{1.90}             & 675M                   & 401                    \\
                                                                           & 2                                  & \textbf{1.88}             & 1.5B                   & 921                    & $\pi$-Flow~\citep{chen2025pi}                             & 1                              & 3.34                      & 675M                   & 448                    \\
            UCGM-XL~\citep{sun2025unified}                                 & 1                                  & 2.63                      & 675M                   & 360                    & $\pi$-Flow\rp REPA-XL/2                                   & 1                              & 2.85                      & 675M                   & 448                    \\
            \midrule \multicolumn{10}{c}{\textbf{Single-branch one-step models (teacher-free)}}                                                                                                                                                                                                                                                                          \\
            \midrule sCT-M~\citep{lu2024simplifying}                       & 1                                  & 5.84                      & 498M                   & 1837                   & Shortcut-XL/2~\citep{frans2024one}                        & 1                              & 10.6                      & 676M                   & 250                    \\
                                                                           & 2                                  & 5.53                      & 498M                   & 1837                   &                                                           & 4                              & 7.80                      & 676M                   & 250                    \\
            sCT-L~\citep{lu2024simplifying}                                & 1                                  & 5.15                      & 778M                   & 1274                   & IMM-XL/2~\citep{zhou2025inductive}                        & 1$\times$2                     & 7.77                      & 675M                   & 3840                   \\
                                                                           & 2                                  & 4.65                      & 778M                   & 1274                   &                                                           & 2$\times$2                     & 5.33                      & 675M                   & 3840                   \\
            sCT-XXL~\citep{lu2024simplifying}                              & 1                                  & 4.29                      & 1.5B                   & 762                    & IMM-XL/2 ($\omega=1.5$)                                   & 1$\times$2                     & 8.05                      & 675M                   & 3840                   \\
                                                                           & 2                                  & 3.76                      & 1.5B                   & 762                    &                                                           & 2$\times$2                     & 3.99                      & 675M                   & 3840                   \\
            MeanFlow-XL/4~\citep{geng2025mean}                             & 1                                  & 4.21                      & 676M                   & 360                    & MeanFlow-XL/2~\citep{geng2025mean}                        & 1                              & 3.43                      & 676M                   & 240                    \\
                                                                           & 2                                  & 3.41                      & 676M                   & 360                    &                                                           & 2                              & 2.93                      & 676M                   & 240                    \\
            MeanFlow-XL/2\rp DC-AE                                         & 2                                  & \textbf{2.73}             & 676M                   & 840                    & MeanFlow-XL/2 (longer training)                           & 2                              & \textbf{2.20}             & 676M                   & 1000                   \\
            \midrule \multicolumn{10}{c}{\textbf{Dual-branch one-step models (teacher-free)}}                                                                                                                                                                                                                                                                            \\
            \midrule \method-XL/4                                          & 1                                  & 3.47                      & 679M                   & 360                    & \method-XL/2                                              & 1                              & 3.07                      & 679M                   & 240                    \\
            \method-XL/4                                                   & 2                                  & 2.76                      & 679M                   & 360                    & \method-XL/2                                              & 2                              & 2.82                      & 679M                   & 240                    \\
            \method-XL/2\rp DC-AE                                          & 1                                  & 2.89                      & 679M                   & 840                    & \method-XL/2 (longer training)                            & 1                              & 2.60                      & 679M                   & 432                    \\
            \method-XL/2\rp DC-AE                                          & 2                                  & \textbf{2.23}             & 679M                   & 840                    & \method-XL/2 (longer training)                            & 2                              & \textbf{1.79}             & 679M                   & 432                    \\
            \bottomrule
        \end{tabular}%
    }
    \vspace{-1em}
\end{table*}

\subsection{Comparison with SOTA Few-step Methods}
\label{sec:SOTA_few}

As presented in \tabref{tab:in512and256_fewsteps}, we primarily compare our \method with \textbf{teacher-free single-branch models}.
Unlike distillation-based approaches that rely on pre-trained teacher networks, these methods (including ours) learn the generation trajectory directly.
We demonstrate that our dual-branch design significantly enhances the efficiency and quality of this direct training paradigm.
\paragraph{Comparison on $512\times512$ resolution.} Under the standard training schedule (360 epochs), our dual-branch paradigm proves more effective than the conventional single-branch design.
We achieve an FID of \textbf{2.76}, surpassing MeanFlow-XL/4 (3.41 FID) by a clear margin.
Notably, even when compared to the parameter-heavy sCT-XXL (1.5B params), our method yields superior results while being significantly more compact.
When equipped with the DC-AE, our performance improves to \textbf{2.23} FID.
This result is particularly impressive as it outperforms sCD-M (2.26 FID)—a distillation-based model—demonstrating that our teacher-free approach can rival the quality of distilled models without their associated pipeline complexity.

\paragraph{Comparison on $256\times256$ resolution.} The superiority of our dual-branch paradigm is further pronounced in the long-training regime.
\method achieves FID \textbf{1.79} with 2 NFEs, establishing a new SOTA for teacher-free models.
We substantially outperform single-branch baselines such as MeanFlow-XL/2 (2.20 FID) and the computationally demanding IMM-XL/2 (3.99 FID, 3840 epochs).
This indicates that the bottleneck in direct consistency training lies not in the training duration, but in the paradigm itself, which our dual-branch formulation effectively resolves.

\textit{%
    \textbf{Summary.} \method sets a new benchmark among Teacher-free One-step Models.
    By addressing the limitations of single-branch paradigm, it consistently outperforms strong baselines like MeanFlow and sCT, offering the best trade-off between generation quality and training simplicity.
}

\subsection{Exploration of Higher-Order Multiplicity}
A natural question arises: \emph{given the efficacy of dual-branch learning, does extending the framework to predict additional geometric quantities (i.e., scaling beyond
    two output heads) yield further gains?}
We investigate this ``Multiplicity Model'' hypothesis on ImageNet-1K $256\times256$ by equipping the backbone with \textbf{three} output heads.
We evaluate two representative configurations trained for 424 epochs:
\begin{enumerate}[label=(\alph{*}), nosep, leftmargin=16pt]
    \item \textbf{Tri-head (FM + MeanFlow + Consistency):}
          Simultaneously predicts velocity, the MeanFlow target~\citep{geng2025mean}, and the consistency mapping.
          The 1-NFE FID degrades to $2.84$ (vs. $2.60$ FID for our dual-branch baseline).

    \item \textbf{Tri-head (FM + Shortcut + Consistency):} Predicts velocity, the Shortcut target~\citep{frans2024one}, and consistency.
          Performance drops to an FID of $4.73$.
\end{enumerate}

We attribute these negative results to two primary factors:
\begin{enumerate}[label=(\alph{*}), nosep, leftmargin=16pt]
    \item \textbf{Homologous redundancy:} Theoretically, Flow Matching, MeanFlow, and Shortcut are all derived from the same underlying PF-ODE mechanics\cite{sun2025unified,sun2026anystep}.
          Consequently, they provide highly correlated supervision signals.
          Adding a third branch offers marginal information gain while merely increasing the optimization complexity.

    \item \textbf{Quality of supervision:} As observed in \tabref{tab:in512and256_fewsteps}, baselines like Shortcut exhibit relatively poor intrinsic performance (FID $>10.0$). Including such objectives acts as a ``noisy teacher'', introducing suboptimal gradients that effectively drag down the overall training trajectory.
\end{enumerate}

\textbf{\textit{Conclusion:}}
The dual-branch design appears to be the optimal ``sweet spot'' within the PF-ODE paradigm.
To achieve further breakthroughs, we conjecture that future work must look beyond homologous ODE solvers and explore \textbf{heterogeneous} algorithms—such as adversarial learning (GANs)—that can provide orthogonal training signals.

\subsection{Ablation Studies}
\label{sec:ablation}

\begin{figure*}[t]
    \centering
    \begin{subfigure}
        [b]{0.325\textwidth}
        \centering
        \includegraphics[width=\linewidth]{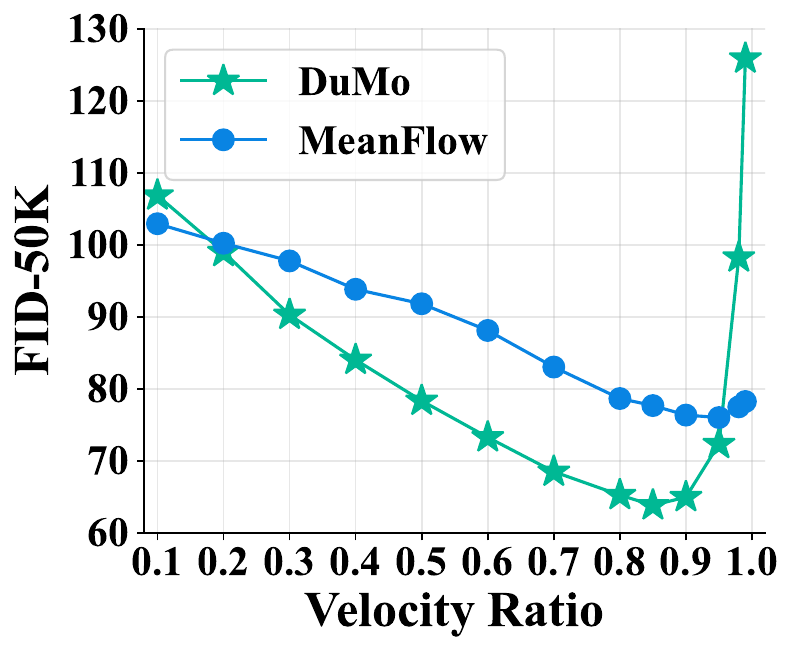}
        \caption{\textbf{Performance vs. velocity ratio ($\beta$).}}
        \label{fig:diff_vel_ratio}
    \end{subfigure}
    \hfill
    \begin{subfigure}
        [b]{0.325\textwidth}
        \centering
        \includegraphics[width=\linewidth]{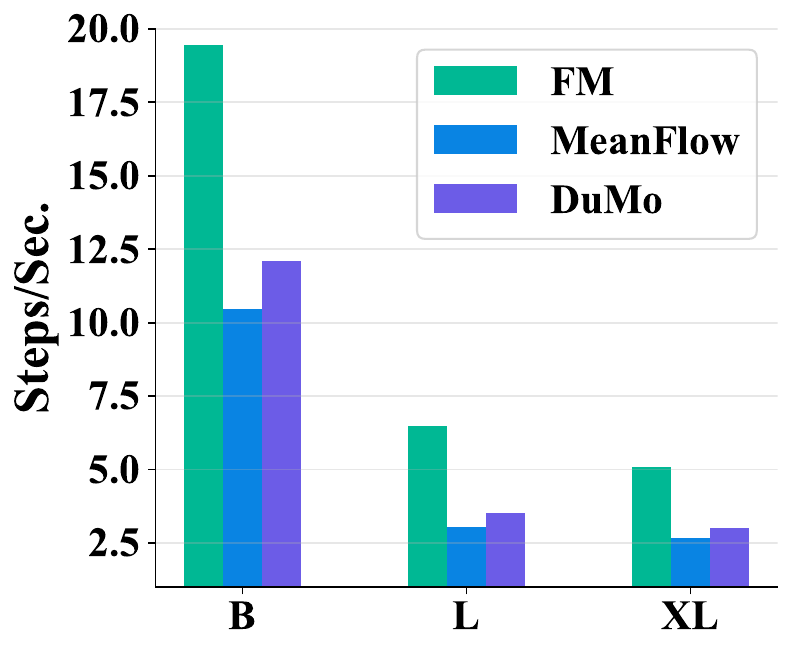}
        \caption{\textbf{Training speed vs. model size.}}
        \label{fig:diff_model_sizes}
    \end{subfigure}
    \hfill
    \begin{subfigure}
        [b]{0.325\textwidth}
        \centering
        \includegraphics[width=\linewidth]{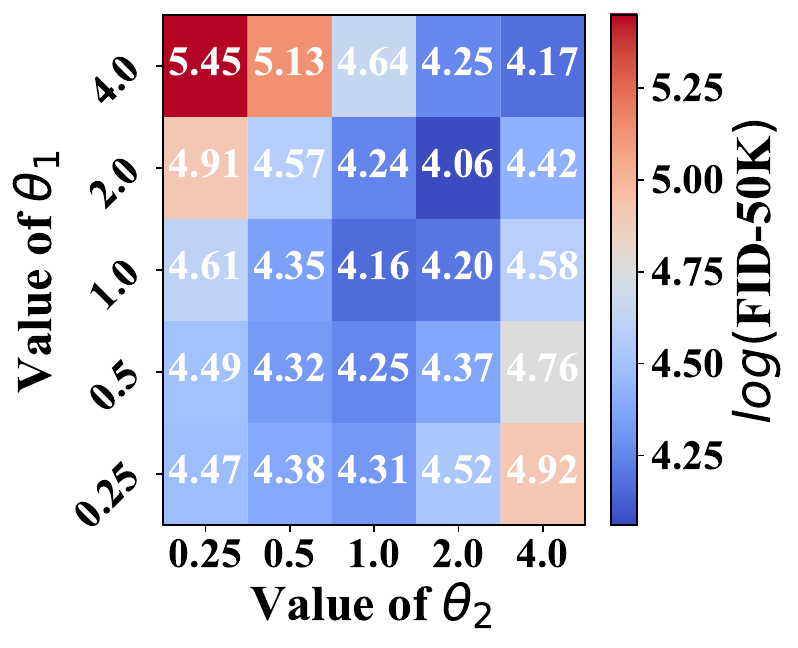}
        \caption{\textbf{Performance sensitivity to $(\theta_{1},\theta_{2})$.}}
        \label{fig:time_dist}
    \end{subfigure}
    \caption{\small{ \textbf{Ablation studies of \method on ImageNet-1K $256\!\times\!256$.} We investigate key design factors of \method for one-step generation, benchmarking against the single-branch baseline, MeanFlow. Unless noted otherwise, experiments in (a) and (c) employ a DiT-B/2 backbone trained for 10K iterations with a learning rate of $2 \times 10^{-4}$, adhering to the optimization protocol in \secref{sec:expset}. }}
    \label{fig:ablation}
    \vspace{-1em}
\end{figure*}

To validate the effectiveness of our design choices, we conduct comprehensive ablation studies on ImageNet $256\times256$.
We analyze three critical dimensions: the sensitivity to the loss weighting factor $\beta$, training computational efficiency, and the impact of time step sampling distributions. The results are summarized in \figref{fig:ablation}.

\paragraph{Impact of velocity ratio ($\beta$).}
\figref{fig:diff_vel_ratio} reveals an advantage for our paradigm.\method (green stars) consistently outperforms the single-branch MeanFlow (blue circles) across the majority of the spectrum, particularly in the effective convergence regimes. Crucially, \method achieves a significantly deeper basin of convergence (best FID=$63.86$) compared to the optimal MeanFlow (best FID=$76.04$).

This empirical observation aligns with our expectation of higher data efficiency: by leveraging unified supervision without strictly partitioning the budget, \method accelerates optimization. This also corroborates the findings in \tabref{tab:in512and256_fewsteps}, explaining why our method yields superior performance under comparable or reduced training steps.

\paragraph{Training efficiency.}
A common concern is that dual-output architectures might incur significant computational overhead.
However, \figref{fig:diff_model_sizes} dispels this assumption.
\method (purple bars) consistently demonstrates higher training throughput (Steps/Sec) compared to MeanFlow (blue bars) across all model sizes (B, L, XL).
Specifically, on the DiT-XL backbone, \method achieves a speedup of approximately $12.6\%$.
This efficiency stems from the streamlined target construction described in \algref{alg:dumo}: by reusing the stop-gradient velocity estimate $\vv_{\mtheta,t}$, we reduce the per-step cost to just 1 gradient-based and 3 gradient-free forward passes (versus 4 for MeanFlow).
Thus, \method balances advanced generative capabilities with computational economy.

\paragraph{Time distribution strategy.}
\figref{fig:time_dist} presents a grid search over the Beta distribution parameters $t \sim \text{Beta}(\theta_{1}, \theta_{2})$, measuring performance via log(FID).
We observe a clear ``sweet spot'' at $\theta_{1}=2.0, \theta_{2}=2.0$, which yields the lowest log-FID of $4.06$ (deepest blue cell).
This corresponds to a symmetric, bell-shaped distribution that concentrates sampling probability in the middle of the trajectory ($t \approx 0.5$).
This finding suggests that the most critical signal for learning the consistency mapping lies in the intermediate noise regime, aligning with observations in prior literature~\citep{sun2025unified,rombach2022high}.

\paragraph{Comparison on CIFAR-10.}
Finally, we extend our evaluation to unconditional generation on CIFAR-10 ($32 \times 32$) to benchmark against a broader range of methods.
As detailed in \tabref{tab:cifar10}, we compare \method against established few-step baselines including iCT~\citep{song2023improved}, ECT~\citep{geng2024consistency}, and IMM. All methods utilize the same U-Net backbone ($\sim 55$M parameters) operating in pixel space.
\method achieves an impressive one-step FID of $2.86$.
This result outperforms both the direct competitor MeanFlow ($2.92$) and sCT ($2.97$), and remains highly competitive with the SOTA iCT ($2.83$), demonstrating the robustness of our approach across different data modalities.

\subsection{Additional Experiments and Analyses}
\label{sec:further_analyses}

Beyond the analysis in \secref{sec:ablation}, we provide more details of the hyper-parameters used in our main experiments in \appref{app:implementation_details}.
In addition to the standard metric (i.e., FID), we assess \method based on (a) qualitative visualization to demonstrate fine-grained details, and (b) the Inception Score (IS), another widely used metric for evaluating the fidelity of generated samples. Comprehensive results are in \appref{app:add_in1k}.

\paragraph{Exploration of real-world applications.}
We validate the scalability of \method on two demanding tasks: text-to-image generation (\appref{app:t2i}) and unified multimodal modeling (\appref{app:unified_multimodal_results}).
Notably, in text-to-image synthesis, \method establishes a new efficiency benchmark, achieving a GenEval score of $0.82$ with just 2 NFE.
This significantly surpasses the previous SOTA, SANA-Sprint~\citep{chen2025sana} ($0.77$), demonstrating superior performance in computationally constrained regimes.

\begin{table}[t]
    \centering
    \caption{\small{\textbf{System-level quality comparison for one-step generation task on unconditional CIFAR-10.}}}
    \label{tab:cifar10}
    \vspace{-0.5em}
    \resizebox{\linewidth}{!}{
        \begin{tabular}{lcccccc}
            \toprule \textbf{Method}           & iCT           & ECT  & sCT  & IMM  & MeanFlow & \method (ours) \\
            \midrule \textbf{NFE~$\downarrow$} & 1             & 1    & 1    & 1    & 1        & 1              \\
            \textbf{FID~$\downarrow$}          & \textbf{2.83} & 3.60 & 2.97 & 3.20 & 2.92     & 2.86           \\
            \bottomrule
        \end{tabular}
    }
    \vspace{-1em}
\end{table}

\section{Conclusion}

In this work, we introduce \textbf{Duality Models} (\method), a unified framework that fundamentally resolves the "budget partition dilemma" inherent in target-aware generative models by embracing the core philosophy we term "\textbf{Let one Do More!}": unifying distinct capabilities within a single representation.
Specifically, by transitioning from a conditional "one input, one output" switch to a parallel "one input, dual output" paradigm, \method effectively leverages the PF-ODE velocity as a geometric regularizer for the flow-map integral. This architectural innovation enables a $679$M parameter DiT to achieve SOTA performance (FID $1.79$) on ImageNet $256\times256$ in just $2$ steps, successfully bridging the stability of flow matching with the efficiency of consistency models without relying on pre-trained teachers.

%% file: resources/appendix.tex
\newpage
\appendix
\onecolumn

\section{Limitations and Future Work}
Despite these advancements, our current exploration is primarily limited to dual-branch systems involving homologous PF-ODE objectives.
While preliminary experiments suggest that simply stacking additional ODE-based heads (e.g., shortcut or MeanFlow) yields diminishing returns due to signal redundancy, the broader potential of \method lies in its ability to fuse distinct generative paradigms.
A critical and unexplored avenue is the integration of \textit{heterogeneous} objectives.
For instance, combining the geometric stability of flow matching with the perceptual sharpness of adversarial learning (GANs) or the statistical alignment of distribution matching losses (e.g., DMD~\citep{yin2024one} or TwinFlow~\citep{cheng2025twinflow}) could provide orthogonal training signals.
We believe \method serves as a foundational blueprint for such unified, multi-objective architectures, opening new possibilities for pushing the boundaries of one-step generation quality.

\section{Related Work}
\label{sec:related_work}

The landscape of continuous-time generative models has evolved from multi-step integration towards high-fidelity, few-step synthesis.
Our work builds upon this trajectory by addressing the limitations of existing paradigms.
We contextualize our contributions by surveying three dominant research thrusts—multi-step foundations, interval-based consistency, and adversarial refinement—and concluding with their specific challenges in large-scale applications.

\subsection{Foundations: Multi-Step Integration of Instantaneous Fields}

The dominant paradigm in continuous generative modeling, including Denoising Diffusion Models~\citep{ho2020denoising,song2020score,dhariwal2021diffusion} and Flow-Matching~\citep{lipman2022flow}, relies on learning an \emph{instantaneous} velocity field.
These models train a neural network to approximate the local dynamics $\frac{\mathrm{d}\mathbf{x}_t}{\mathrm{d}t}$ of a Probability Flow Ordinary Differential Equation (PF-ODE).
To generate a sample, one must numerically integrate this field, typically requiring hundreds or thousands of steps to ensure fidelity.
The core limitation of this approach is its sensitivity to coarse discretization; when using few steps, large truncation errors accumulate, particularly for trajectories with high curvature, leading to a significant degradation in sample quality~\citep{karras2022elucidating}.
This challenge has catalyzed the development of methods designed specifically for the few-step regime.

\subsection{Interval-Based Consistency for Few-Step Generation}

A major research thrust aims to overcome the integration bottleneck by enforcing consistency over finite time intervals or distilling the trajectory.
Early distillation methods~\citep{Luhman2021KnowledgeDI,salimans2022progressive} trained students to match teacher steps.
Consistency Models (CMs)~\citep{song2023consistency,song2023improved} advanced this by enforcing a \emph{relative} constraint: the model's prediction of the trajectory's endpoint ($\xx_0$) should be consistent across different starting points.
This concept was extended by methods like MeanFlow~\citep{geng2025mean}, which directly models the average velocity.

\paragraph{Improving Training Efficiency and Objectives.}
A critical implementation challenge is the efficient computation of time derivatives.
Early methods relied on Jacobian-Vector Products (JVP)~\citep{geng2025mean,lu2024simplifying}, which are computationally intensive and incompatible with modern optimizations like FlashAttention~\citep{dao2022flashattention} and FSDP~\citep{zhao2023pytorch}.
Recent work addresses this bottleneck: \citet{sun2025unified} propose finite-difference estimators, while SoFlow~\citep{luo2025soflow} introduces a solution consistency loss that bypasses JVP entirely.
Similarly, T2I-Distill~\citep{pu2025few} develops efficient kernels for open-domain generation.
New approaches continue to emerge: FACM~\citep{peng2025facm} stabilizes training by anchoring to the instantaneous field, $\pi$-Flow~\citep{chen2025pi} adopts an imitation learning perspective, and Self-E~\citep{yu2025self} leverages self-evaluation to refine generation.

\subsection{Adversarial and Distribution-Matching Refinement}

A parallel approach achieves high-fidelity, one-step generation by incorporating external signals—either adversarial or distribution-matching—to explicitly anchor outputs to the data manifold.
Methods such as CTM~\citep{kim2023consistency}, ADD/LADD~\citep{sauer2024adversarial,sauer2024fast}, and distillation techniques like DMD/DMD2~\citep{yin2024one,yin2024improved} employ auxiliary discriminators or minimizing KL divergence to sharpen model outputs.
Notably, recent works like TwinFlow~\citep{cheng2025twinflow} further advance this direction by coupling dual flows to enforce distribution consistency without relying heavily on unstable adversarial min-max games.
Other GAN-based refiners~\citep{zheng2025direct} similarly push the student to surpass the teacher.

Recent advancements refine this paradigm: Decoupled-DMD~\citep{liu2025decoupled} identifies CFG as the primary acceleration driver, while DMD-R~\citep{jiang2025distribution} integrates reinforcement learning for better mode coverage.
However, this reliance is a double-edged sword.
Adversarial and distribution-matching frameworks typically depend on a frozen, pre-trained teacher to generate synthetic data, the cost of which can be prohibitive for ultra-large models~\citep{yin2024improved}.
Moreover, auxiliary networks increase memory overhead and training instability.
Approaches like \citet{tong2025flow} attempt to mitigate this via data-free distillation from prior distributions, but balancing fidelity and complexity remains a key challenge.

\subsection{Applications in Large-Scale Generative Models}

The tension between speed and quality is amplified in state-of-the-art large-scale systems.
For instance, Qwen-Image-20B~\citep{wu2025qwenimagetechnicalreport} typically requires 100 NFEs, resulting in substantial latency ($\sim$40s for a 1024$\times$1024 image on a single A100).
While pipelines like Hunyuan-Image-2.1~\citep{HunyuanImage-2.1} have explored MeanFlow for mid-step acceleration, and others apply LCM-style or hybrid distillation (e.g., PixArt-Delta~\citep{chen2024pixartdelta}, SDXL~\citep{lin2024sdxl}, FLUX-schnell~\citep{flux2024}, SANA-Sprint~\citep{chen2025sana}, LCM~\citep{luo2023latent}), targeting extreme acceleration (1-2 NFEs) for 10B-20B backbones remains difficult.
The prohibitive memory costs of teacher-dependent pipelines and the instability of adversarial training hinder straightforward scaling.
Beyond static images, this challenge extends to video generation, where recent works like rCM~\citep{zheng2025large} and TMD~\citep{nie2026transition} represent initial steps towards applying continuous-time consistency and decoupled distillation to temporally complex data.

\section{Detailed Experiment}
\label{app:exp}

\subsection{Implementation Details}
\label{app:implementation_details}

\paragraph{Training setup.}
We implement all models using PyTorch and train them on NVIDIA GPUs.
We employ the AdamW optimizer~\cite{loshchilov2017decoupled} with a global batch size of 512.
The learning rate follows a constant schedule throughout training, with no warmup phase.
To stabilize sampling, we apply Exponential Moving Average (EMA) to the model parameters with a decay rate of 0.9999.

\paragraph{Hyperparameter configuration.}
Detailed hyperparameter settings for ImageNet-1K training are provided in \tabref{tab:hyper_param}.
Regarding the specific coefficients for \method :
the Beta distribution parameters $(\theta_1, \theta_2)$ determine the shape of the time sampling distribution during training (illustrated in \figref{fig:time_dist});
the parameter $\beta$ modulates the velocity ratio, representing the weight of the multi-step loss (see \figref{fig:diff_vel_ratio});
and $\zeta$ governs the scale of the training-time classifier-free guidance, following \citet{sun2025unified}.

\begin{table}[h]
    \centering
    \caption{\small \textbf{Hyperparameter configurations for \method on ImageNet-1K.} This table details the settings for few-step training and sampling across different resolutions and architectures.}
    \label{tab:hyper_param}
    \resizebox{\textwidth}{!}{
        \begin{tabular}{@{}cccccccccc@{}}
            \toprule
            \multicolumn{3}{c|}{Task} & \multicolumn{3}{c|}{Optimizer} & \multicolumn{4}{c}{\method}                                                                                                                                                                                                                     \\ \midrule
            Resolution                & VAE/AE                         & \multicolumn{1}{c|}{Model}  & Type  & lr     & \multicolumn{1}{c|}{($\beta_1$,$\beta_2$)} & Transport & ($\theta_1$,$\theta_2$) (see \figref{fig:time_dist}) & $\beta$ (see \figref{fig:diff_vel_ratio}) & $\zeta$ (see \citet{sun2025unified}) \\ \midrule
            \multicolumn{10}{c}{Few-step model training and sampling}                                                                                                                                                                                                                                                    \\ \midrule
            256                       & SD-VAE                         & \multicolumn{1}{c|}{XL/2}   & AdamW & 0.0001 & \multicolumn{1}{c|}{(0.9,0.95)}            & Linear    & (2.4,2.4)                                            & 0.7                                       & 0.55                                 \\ \midrule
            512                       & DC-AE                          & \multicolumn{1}{c|}{XL/1}   & AdamW & 0.0001 & \multicolumn{1}{c|}{(0.9,0.95)}            & Linear    & (1.0,1.0)                                            & 0.7                                       & 0.66                                 \\
                                      & SD-VAE                         & \multicolumn{1}{c|}{XL/4}   & AdamW & 0.0001 & \multicolumn{1}{c|}{(0.9,0.95)}            & Linear    & (2.4,2.4)                                            & 0.7                                       & 0.66                                 \\ \bottomrule
        \end{tabular}}
    \vspace{-1em}
\end{table}

\subsection{Additional Results on ImageNet-1K}
\label{app:add_in1k}

In this section, we provide a comprehensive evaluation of generation quality on ImageNet-1K ($256\times256$), complementing the main results in \secref{sec:exp}.
Beyond the standard Fr\'echet Inception Distance (FID), we report the Inception Score (IS)~\citep{salimans2016improved} to explicitly assess the diversity and class-conditional distinctiveness of the generated samples.
We compare our proposed \method against the strongest single-branch baseline, MeanFlow~\citep{geng2025mean}, under identical training budgets ($432$ epochs).

\begin{figure*}[t!]
    \centering
    \includegraphics[width=\linewidth]{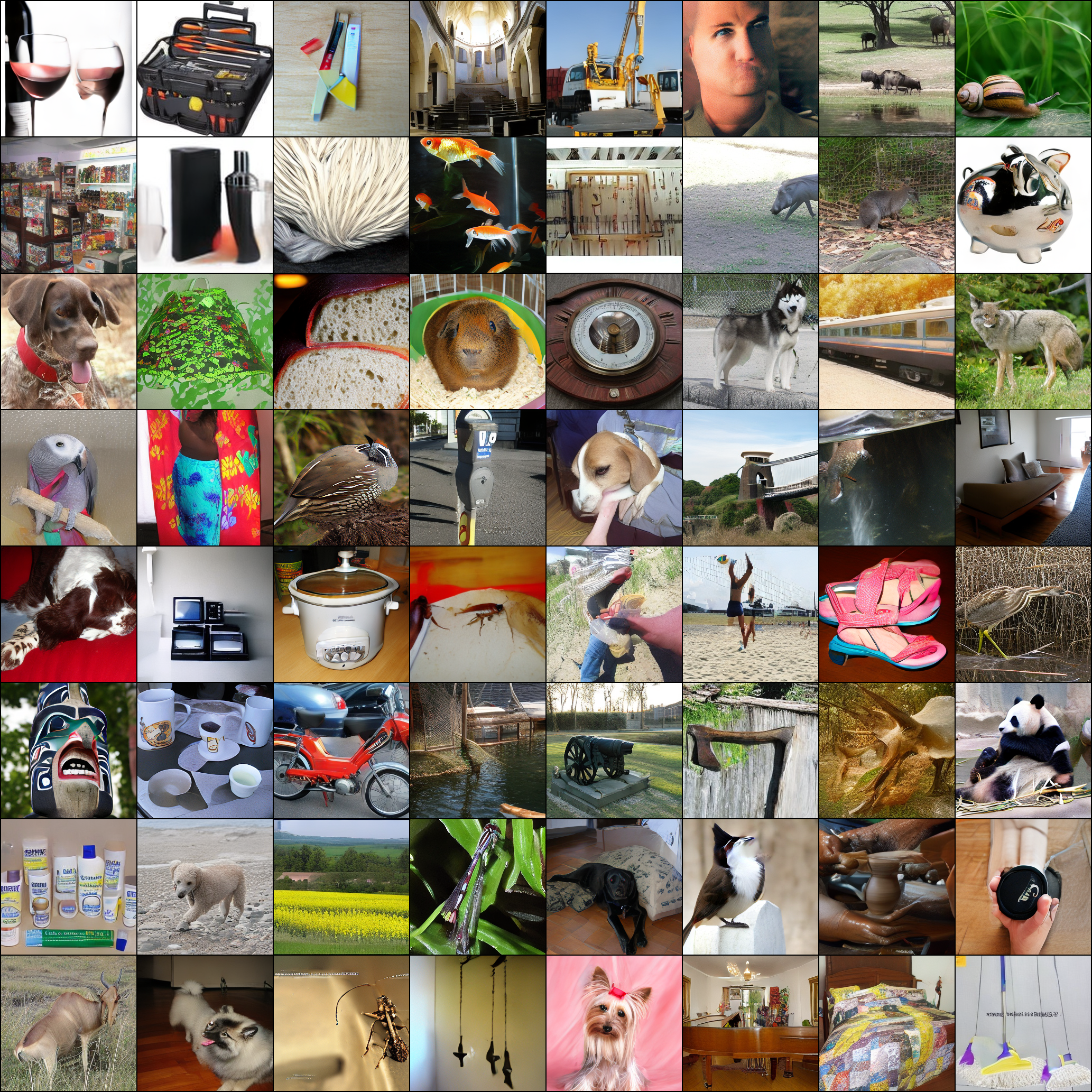}
    \caption{\small{
            \textbf{Visualization of randomly generated images ($256\times256$) from MeanFlow-XL/2~\citep{geng2025mean}. The model was trained for 432 epochs and sampled using 1 NFE, achieving FID$=2.87$ and IS$=239.58$.}
        }}
    \label{fig:meanflow_1nfe}
\end{figure*}

\begin{figure*}[t!]
    \centering
    \includegraphics[width=\linewidth]{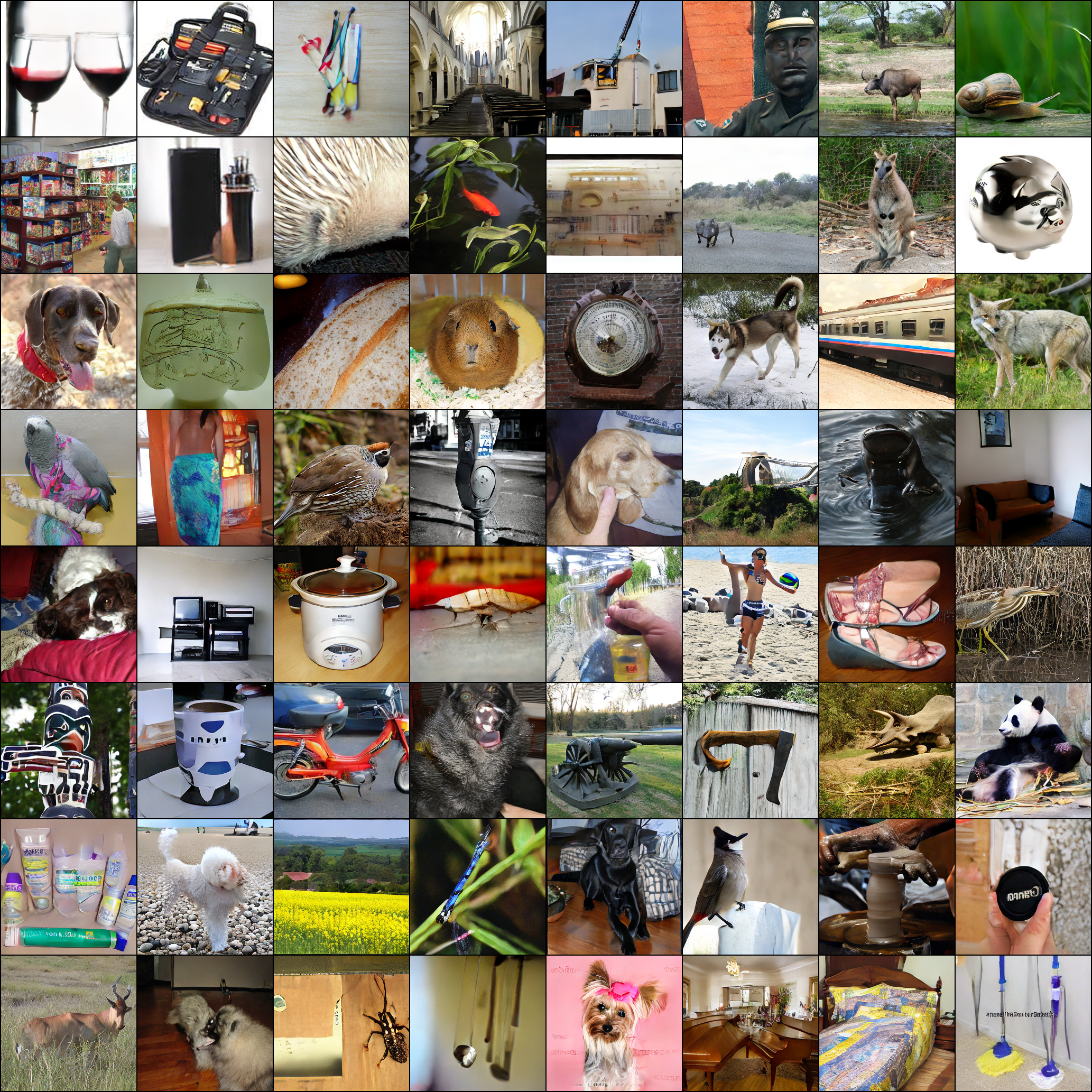}
    \caption{\small{
            \textbf{Visualization of randomly generated images ($256\times256$) from our \method-XL/2. The model was trained for 432 epochs and sampled using 1 NFE, achieving FID$=2.60$ and IS$=246.26$.}
        }}
    \label{fig:ours_1nfe} %
\end{figure*}

\begin{figure*}[t!]
    \centering
    \includegraphics[width=\linewidth]{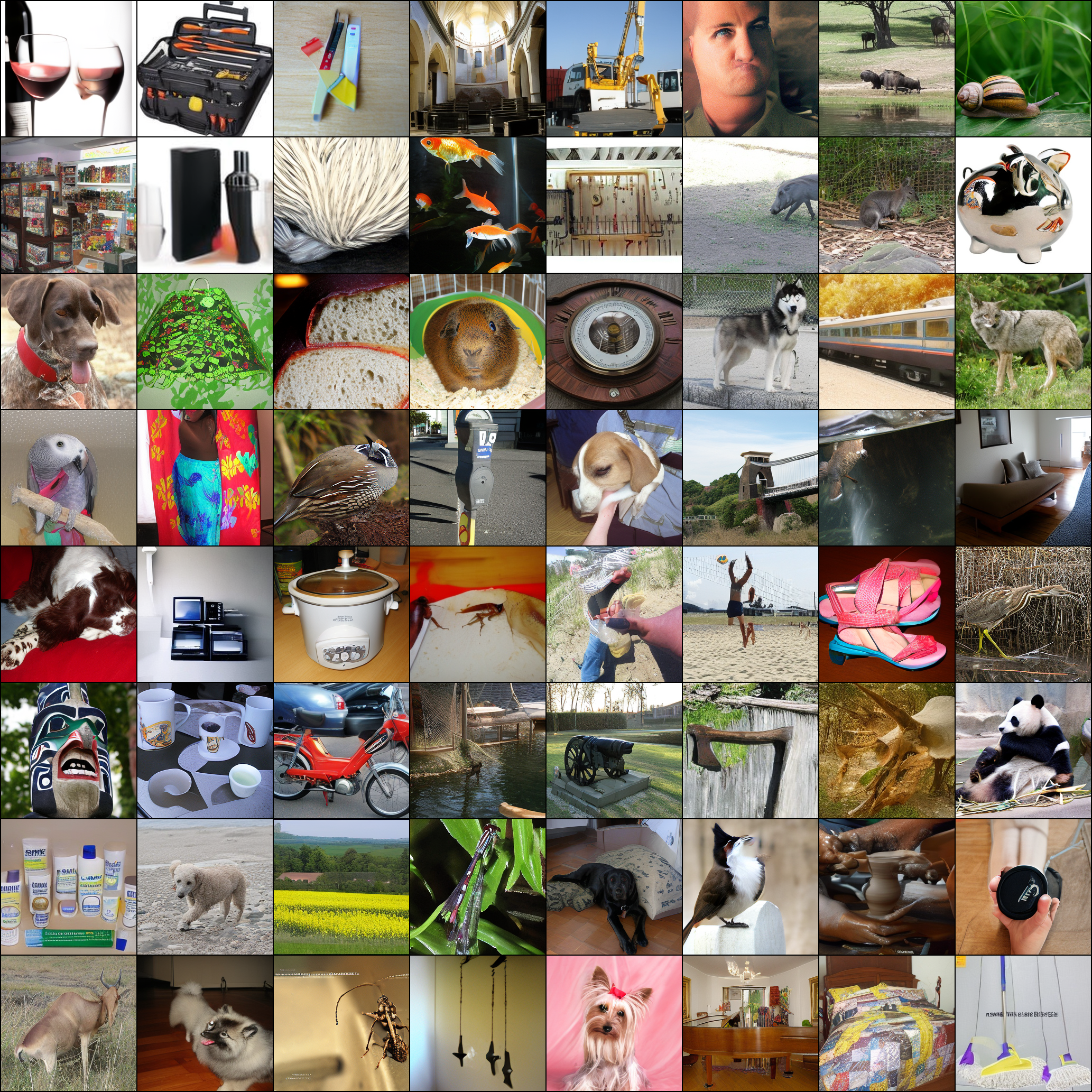}
    \caption{\small{
            \textbf{Visualization of randomly generated images ($256\times256$) from MeanFlow-XL/2~\citep{geng2025mean}. The model was trained for 432 epochs and sampled using 2 NFE, achieving FID$=2.39$ and IS$=254.39$.}
        }}
    \label{fig:meanflow_2nfe} %
\end{figure*}

\begin{figure*}[t!]
    \centering
    \includegraphics[width=\linewidth]{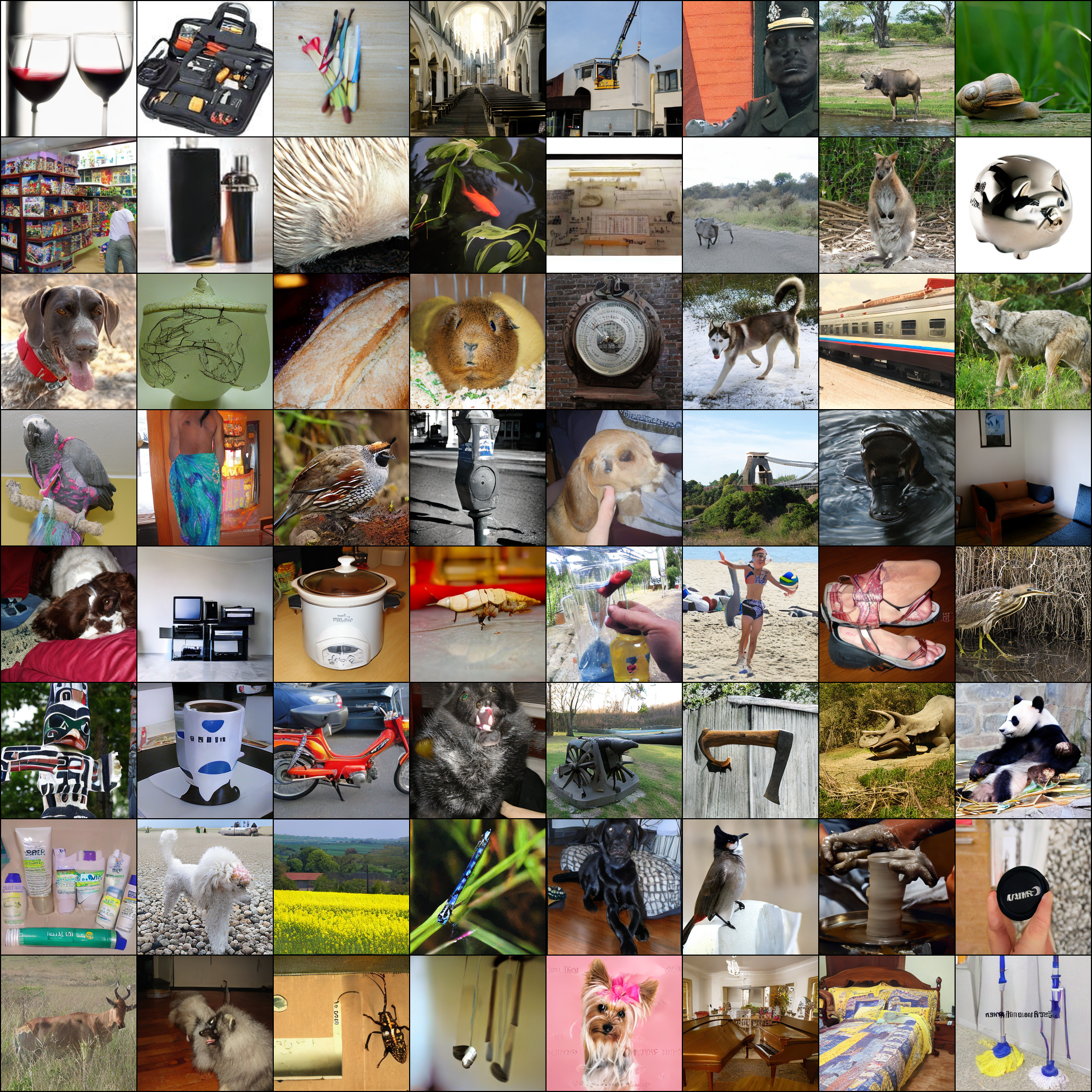}
    \caption{\small{
            \textbf{Visualization of randomly generated images ($256\times256$) from our \method-XL/2. The model was trained for 432 epochs and sampled using 2 NFE, achieving FID$=1.79$ and IS$=285.67$.}
        }}
    \label{fig:ours_2nfe} %
\end{figure*}

\paragraph{Quantitative analysis.}
As illustrated in \figref{fig:meanflow_1nfe} through \figref{fig:ours_2nfe}, \method consistently outperforms MeanFlow across both sampling regimes.
\begin{enumerate}[label=(\alph{*}), nosep, leftmargin=16pt]
    \item \textbf{One-step Generation:}
          At $1$ NFE, \method achieves an IS of $\mathbf{246.26}$ and an FID of $\mathbf{2.60}$, surpassing MeanFlow (IS $239.58$, FID $2.87$).
          This confirms that our dual-branch design effectively mitigates the ``starvation'' of the generative head, allowing for better convergence even in the ultra-low step regime.

    \item \textbf{Two-step Generation:}
          The performance gap widens significantly when a second denoising step is allowed.
          \method achieves a remarkable IS of $\mathbf{285.67}$ and an SOTA FID of $\mathbf{1.79}$ (\figref{fig:ours_2nfe}).
          In contrast, MeanFlow saturates at an IS of $254.39$ and FID of $2.39$ (\figref{fig:meanflow_2nfe}).
          The substantially higher Inception Score ($+31.28$ points) indicates that our model captures a much richer diversity of ImageNet classes with higher confidence, validating the efficacy of the velocity-guided regularization.
\end{enumerate}

\paragraph{Qualitative analysis.}
We provide randomly sampled visualizations to corroborate these metrics.
\figref{fig:meanflow_1nfe} and \figref{fig:meanflow_2nfe} show samples from MeanFlow. While structurally coherent, they occasionally exhibit texture over-smoothing or minor artifacts in complex geometries.
In comparison, samples from \method (\figref{fig:ours_1nfe} and \figref{fig:ours_2nfe}) demonstrate superior perceptual fidelity.
Notably, in the $2$-step regime, our model generates images with sharp fine-grained details and accurate semantic structures, closely approximating the data distribution.
This visual evidence aligns with our hypothesis: by simultaneously enforcing local geometric constraints (velocity) and global consistency (flow-map) on every sample, \method learns a more robust and accurate transport trajectory.

\subsection{Comparison with Text-to-image Models}
\label{app:t2i}

To investigate the versatility of the \method paradigm beyond the scope of standard class-conditional generation, we extend our evaluation to the more challenging text-to-image synthesis task.
This benchmark serves to verify whether the architectural advantages of our dual-branch design can effectively translate to open-domain scenarios involving complex semantic control.
Comprehensive experimental analyses and qualitative comparisons will be presented in the forthcoming version of this manuscript.

\subsection{Experimental Results on Large-scale Unified Multimodal Models}
\label{app:unified_multimodal_results}

To rigorously assess the robustness and scalability of \method under regimes of extreme parameter counts, we further evaluate our framework within the domain of Unified Multimodal Models (UMMs).
We benchmark our approach against state-of-the-art large-scale foundational models, such as Qwen-Image-20B~\citep{wu2025qwenimagetechnicalreport} and OpenUni~\citep{wu2025openuni}, to validate the efficacy of the proposed objective at scale.
The detailed quantitative results from this large-scale evaluation will be updated in the future manuscript.

\begin{figure}[h]
    \centering
    \includegraphics[width=\linewidth]{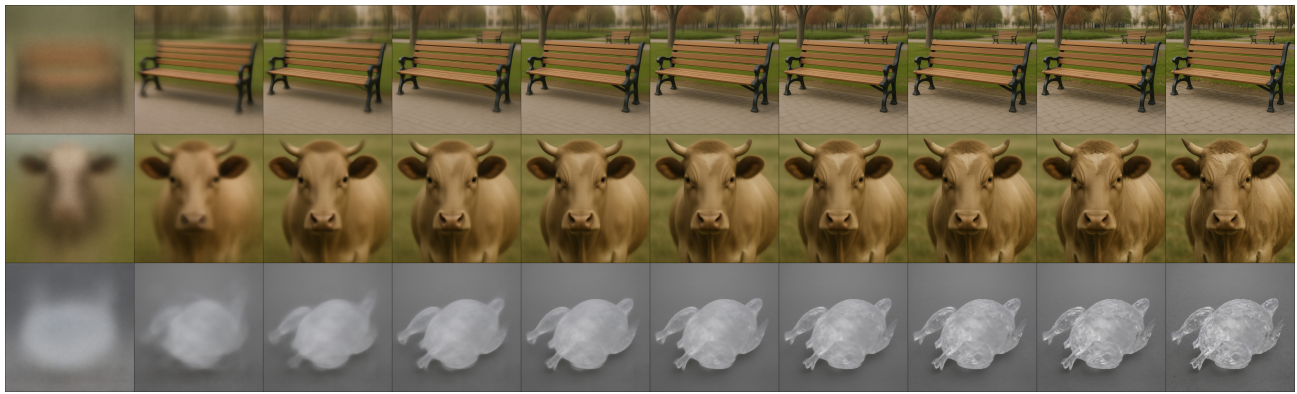}
    \vspace{-1.5em}
    \caption{\small
        \textbf{Visualization of sampling trajectories at the baseline state (0 training steps).}
        Initially, the model requires significant computational steps to resolve coherent images. The leftmost columns (low NFE) remain unstructured and blurry, with high-fidelity results only emerging in the rightmost columns, indicating a highly curved generation trajectory typical of the teacher model.
    }
    \label{fig:viz_0}
    \vspace{-1em}
\end{figure}

\begin{figure}[h]
    \centering
    \includegraphics[width=\linewidth]{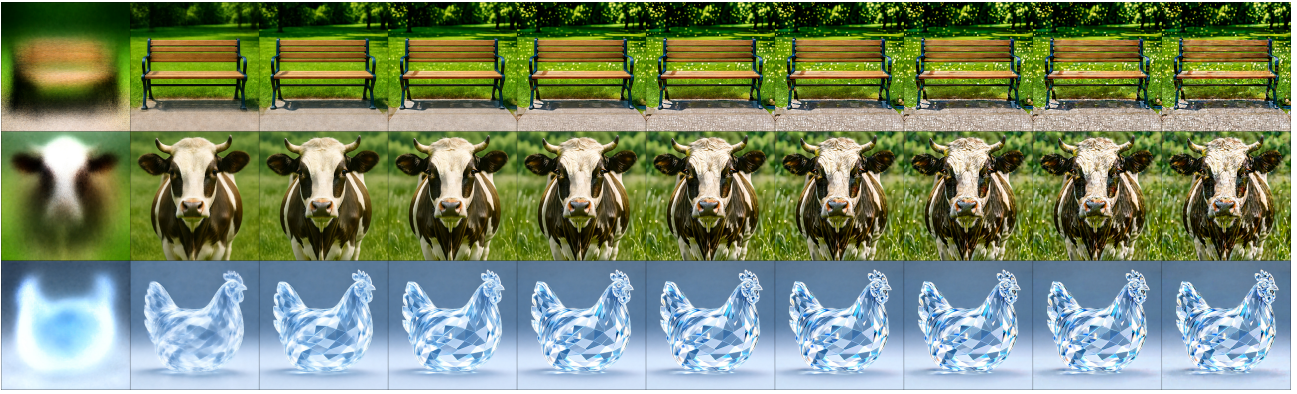}
    \vspace{-1.5em}
    \caption{\small
        \textbf{Visualization of sampling trajectories demonstrating the trend towards convergence (3000 training steps).}
        At this intermediate training stage, \method already shows significant improvement in few-step generation. Compared to~\figref{fig:viz_0}, clear object structures emerge much earlier at lower NFEs (left columns), visually affirming that the generation trajectory is being effectively straightened towards a few-step model.
    }
    \label{fig:viz_3000}
    \vspace{-1em}
\end{figure}

\paragraph{Qualitative analysis of convergence trend.}
To explicitly verify the convergence trend of the model towards a few-step generator, we compare the visualization of sampling trajectories at different training stages.
As shown in~\figref{fig:viz_0} (0 steps), the initial model follows the teacher's curved trajectory, where coherent image structure only emerges at higher NFEs (right columns), with low-NFE outputs remaining largely unstructured.
Conversely,~\figref{fig:viz_3000} (3000 steps) clearly demonstrates the model is converging: even at the lowest NFEs (leftmost columns), the generated images—such as the park bench, cow, and crystal chicken—already exhibit clear structures and recognizable details, representing a significant improvement over the initial state.
This visual evidence confirms that \method is effectively straightening the generation flow, progressively enabling the model to map noise to high-quality data in fewer steps.

\section{Theoretical Analysis}
\label{app:theoretical_analysis}

\paragraph{Intrinsic encompassment of multi-step dynamics in consistency learning.}

To understand the theoretical legitimacy of the proposed \method, we analyze the composition of the consistency training objective. Following the unified perspective established in \citet{sun2025unified}, we demonstrate that the learning target of one-step consistency models inherently encompasses the objective of multi-step flow matching.

\begin{theorem}[Surrogate objective for unified linear case ($\alpha(t) = t, \ \gamma(t) = 1-t$), see \citet{sun2025unified}]
    \label{thm:surrogate_objective_linear}
    The training of consistency-based models can be analyzed via a surrogate loss function $\mathcal{L}(\mtheta, \lambda)$.
    Here, $\lambda \in (0, 1)$ denotes the \textbf{Consistency Ratio}.
    This objective asymptotically recovers flow matching models as $\lambda \to 0$ and approximates consistency models as $\lambda \to 1$.
    Crucially, for any $\lambda$, the objective decomposes as:
    \begin{align}
        \mathcal{L}(\mtheta, \lambda)
         & = \E_{\zz, \xx,\, t}
        \Bigg[
            \underbrace{\big\lVert \mmF_{\mtheta}(\xx_t, t) - \zz_t \big\rVert_2^2}_{\text{Flow Matching Term}} + \frac{\lambda}{1 - \lambda} \,
            \underbrace{\big\lVert \mmF_{\mtheta}(\xx_t, t) - \mmF_{\mtheta^{-}}(\xx_{\lambda t}, \lambda t) \big\rVert_2^2}_{\text{Self-Alignment Term}}
            \Bigg],
        \label{eq:surrogate_objective_linear_main}
    \end{align}
    where $\xx_t = t \cdot \zz + (1 - t) \cdot \xx$ and $\zz_t = \zz - \xx$.
\end{theorem}

\thmref{thm:surrogate_objective_linear} provides two critical insights that justify the \method paradigm:

\paragraph{Gradient compatibility.}
\eqref{eq:surrogate_objective_linear_main} reveals that the standard consistency objective is effectively a superposition of a velocity regression loss (first term) and a self-alignment loss (second term).
Consequently, the gradient direction derived from consistency training \textit{already contains} the gradient component required for flow matching.
This theoretical inclusion implies that explicitly adding a flow matching loss via a secondary head (as done in \method) does not introduce adversarial gradients or objective conflicts. Instead, it acts as a \textbf{compatible reinforcement}, explicitly supervising a signal that the consistency objective was already implicitly attempting to minimize.

\paragraph{Implicit-to-explicit duality.}
Since the flow matching term is an intrinsic component of the surrogate objective, a well-trained one-step consistency model must, by definition, implicitly capture the multi-step velocity field to achieve minimal loss.
In other words, a functional one-step generator naturally contains a multi-step estimator within its latent representations.
Therefore, the design of \method—extracting this implicit knowledge via a dedicated velocity head—is a natural architectural evolution. It allows the shared backbone to learn a unified representation that is valid for both local differential dynamics ($\vv_t$) and global integral mappings ($\uu_t$) without representation collapse or task interference.

\newpage

